%% file: main.tex
\documentclass[nohyperref]{article}

\usepackage{microtype}
\usepackage{graphicx}
\usepackage{subcaption}
\usepackage{booktabs} % for professional tables

\usepackage{hyperref}

\usepackage[accepted]{icml2022}

\usepackage{amsmath}
\usepackage{amssymb}
\usepackage{mathtools}
\usepackage{amsthm}

\usepackage[capitalize,noabbrev]{cleveref}

\theoremstyle{plain}

\theoremstyle{definition}

\theoremstyle{remark}

\usepackage[textsize=tiny]{todonotes}

\usepackage{booktabs}       % professional-quality tables
\usepackage{amsfonts}       % blackboard math symbols
\usepackage{nicefrac}       % compact symbols for 1/2, etc.
\usepackage{microtype}      % microtypography
\usepackage{xcolor}         % colors
\usepackage{amsmath}
\usepackage{subcaption}
\usepackage{graphicx}
\usepackage{todonotes}
\usepackage{tabularx}
\usepackage{makecell}
\usepackage{enumitem}
\usepackage{color, colortbl}
\usepackage{float}
\usepackage{rotating}
\usepackage{tabularx}
\usepackage{svg}
\usepackage{tablefootnote}
\usepackage{multirow,nicematrix}
\usepackage{cleveref}

\newcommand{\ours}{\textsc{RExC}}

\definecolor{Gray}{HTML}{FFFBCE}
\definecolor{Green}{HTML}{bfffdb}
\definecolor{Red}{HTML}{ffbfbf}

\newcommand{\newpara}[1]{\vspace{0em} \noindent \textbf{#1} \hspace{0.1em}}

\icmltitlerunning{Knowledge-Grounded Self-Rationalization via Extractive and NL Explanations}

\begin{document}

\twocolumn[
% \icmltitle{Self-Rationalization with Commonsense via\\ Extractive and Natural Language Explanations}

% \icmltitle{Self-Rationalization with Knowledge Grounding via\\ Extractive and Natural Language Explanations}

\icmltitle{Knowledge-Grounded Self-Rationalization via\\ Extractive and Natural Language Explanations}

% \icmltitle{Self-Rationalization with Background Knowledge via\\ Extractive and Natural Language Explanations}



\icmlsetsymbol{equal}{*}

\begin{icmlauthorlist}
\icmlauthor{Bodhisattwa Prasad Majumder}{ucsd}
\icmlauthor{Oana-Maria Camburu}{oxford}
\icmlauthor{Thomas Lukasiewicz}{tuw,oxford}
\icmlauthor{Julian McAuley}{ucsd}
\end{icmlauthorlist}

\icmlaffiliation{ucsd}{Department of Computer Science and Engineering, UC San Diego, USA.}
\icmlaffiliation{oxford}{Department of Computer Science, University of Oxford, UK.}
\icmlaffiliation{tuw}{Institute of Logic and Computation, TU Wien, Austria}

\icmlcorrespondingauthor{Bodhisattwa Prasad Majumder}{bmajumde@eng.ucsd.edu}

\icmlkeywords{Machine Learning, ICML}

\vskip 0.3in
]



\printAffiliationsAndNotice{}  % leave blank if no need to mention equal contribution

\begin{abstract}
%An increasing number of works focus on building models that generate 
Models that generate extractive rationales (i.e., subsets of features) or natural language explanations (NLEs) for their predictions are important for explainable AI. While an extractive rationale provides a quick view of the features most responsible for a prediction, an NLE allows for a comprehensive description of the decision-making process behind a prediction. However, current models that generate the best extractive rationales or NLEs often fall behind the state-of-the-art (SOTA) in terms of task performance. In this work, we bridge this gap by introducing \ours{}, a self-rationalizing framework that grounds its predictions and two complementary types of explanations (NLEs and extractive rationales) in background knowledge. Our framework improves over previous methods by: (i) reaching SOTA task performance while also providing explanations, (ii)~providing two types of explanations, while existing models usually provide only one type, and (iii)~beating by a large margin the previous SOTA in terms of quality of both types of explanations. Furthermore, a perturbation analysis in \ours{}~shows a high degree of association between explanations and predictions, a necessary property of faithful explanations.

\end{abstract}

\input{files/intro}
\input{files/method}
\input{files/experiments}

\input{files/results}
\input{files/faithfulness}

\input{files/related_works}

\input{files/conclusion}
% \input{files/table}


\bibliography{example_paper}
\bibliographystyle{icml2022}

%%%%%%%%%%%%%%%%%%%%%%%%%%%%%%%%%%%%%%%%%%%%%%%%%%%%%%%%%%%%%%%%%%%%%%%%%%%%%%%
%%%%%%%%%%%%%%%%%%%%%%%%%%%%%%%%%%%%%%%%%%%%%%%%%%%%%%%%%%%%%%%%%%%%%%%%%%%%%%%
% APPENDIX
%%%%%%%%%%%%%%%%%%%%%%%%%%%%%%%%%%%%%%%%%%%%%%%%%%%%%%%%%%%%%%%%%%%%%%%%%%%%%%%
%%%%%%%%%%%%%%%%%%%%%%%%%%%%%%%%%%%%%%%%%%%%%%%%%%%%%%%%%%%%%%%%%%%%%%%%%%%%%%%
\newpage
\clearpage
\appendix
\input{files/appendix}

\end{document}

%% file: files/intro.tex
\section{Introduction}
% \todo{mention two aspects of NLE: plausibility for justification and faithfulness for reflecting model reasoning so that they can be trusted, no illusion}
%Two of the currently predominant 
Two
approaches 
that currently predominate
for building self-explainable neural models are (i) selecting a subset of input features responsible for a prediction, known as an \emph{extractive rationale} (ER) \cite{DBLP:conf/emnlp/ZaidanE08, DBLP:conf/acl/BastingsAT19, LEI21AAAI}, and (ii) generating a \emph{natural language explanation} (NLE) for a prediction \cite{DBLP:conf/cvpr/ParkHARSDR18,  DBLP:conf/eccv/HendricksARDSD16, DBLP:conf/nips/CamburuRLB18, kayser21}. For an explanation (ER or NLE), one is interested in two characteristics: \textit{quality} (or \textit{plausibility}) and \textit{faithfulness}. Quality measures the degree of matching between the model's explanations and some ground truth; models with low-quality explanations would be undeployable. %, indicating how accurate and actionable they are for practical use.
% A model whose explanations are far from the expected ground-truth would not be of interest for deployment.
\textit{Faithfulness} measures how well the explanations reflect the decision-making processes behind the predictions; unfaithful explanations would be misleading.

ERs are concise and provide quick explanations, which may sometimes be enough for users to assess the trustworthiness of the model. However, ERs may not have the means to provide important details of the reasoning of a model (e.g., relations between features) \citep{DBLP:journals/corr/abs-2010-12762}. %paint the full picture of the decision-making process of the model. % \cite{kaur20, Camburu21WorkshopAAAI, alufaisan2020does}. 
In such cases, NLEs can be complementary, as they allow for detailed justification in a form that is most accessible to humans (natural language). However, machine-generated NLEs, like other generated text, are prone to lacking background knowledge (e.g., commonsense) \cite{camburu2020acl, DBLP:conf/emnlp/MaoMMC19}. This could be because the NLEs are unfaithful or the model did not use the necessary knowledge in its decision-making process. Despite the complementary nature of ERs and NLEs, self-rationalizing models usually provide only one of them, with a few exceptions \cite{DBLP:conf/cvpr/ParkHARSDR18, DBLP:journals/corr/abs-1809-02805}. Moreover, while knowledge grounding has been done for black-box models  \cite{DBLP:conf/emnlp/BauerWB18, DBLP:conf/acl/ChanduBB21, DBLP:conf/emnlp/ChenSYW20}, we are not aware of any work on knowledge grounding for self-rationalizing models. Furthermore, existing self-rationalizing models are often outperformed by black-box models at solving the task at hand, leading to an undesirable performance-explainability trade-off.

\begin{figure*}[t!]
    \centering
    \includegraphics[trim= 187 510 225 160,clip, width=0.95\textwidth]{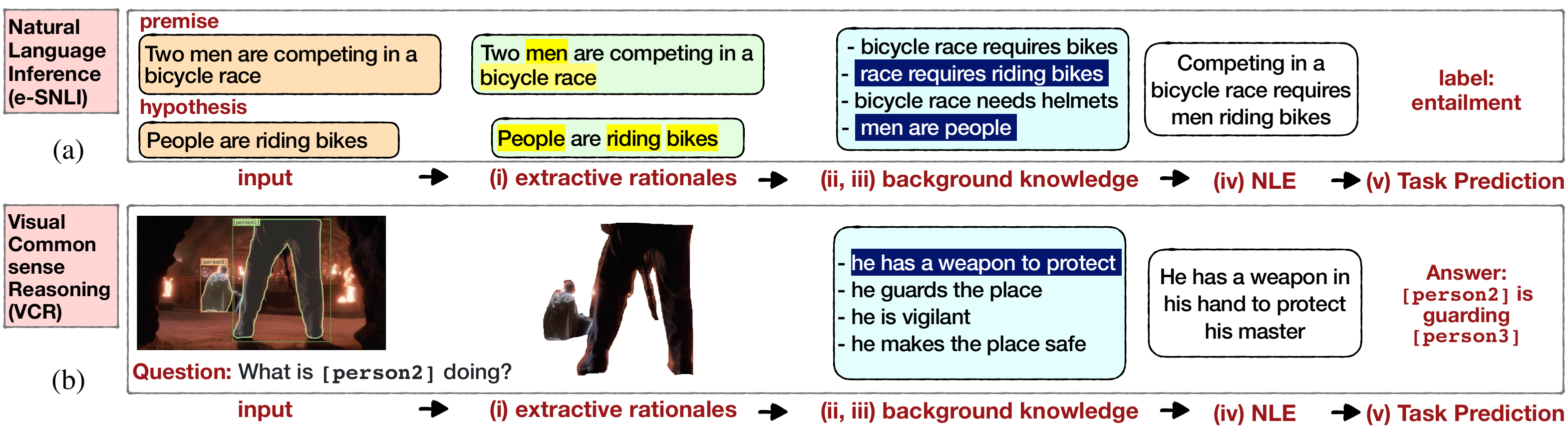}
    \vspace{-0.7em}
    \caption{\small \textbf{Illustrative examples for \ours{}}~on (a) natural language and (b) vision-language tasks.
    % illustrated on two examples.
    }
    \label{fig:overview}
    \vspace{-0.5em}
\end{figure*}

To ground both decision-making and rationalization in background knowledge, as well as to reap the benefits of both ERs and NLEs, we combine these three ingredients in a unified self-rationalization framework. Our framework, which we call \textbf{\ours{}} (Extractive \textbf{R}ationales, Natural Language \textbf{Ex}planations, and (here) \textbf{C}ommonsense)\footnote{Code is available at \url{https://github.com/majumderb/rexc}}, performs five steps: (i) selects a subset of the input features as an ER, (ii) inputs the ER to a knowledge resource to obtain a set of knowledge snippets about the ER, (iii) selects a subset of the snippets as the most relevant ones for solving the instance, (iv) passes the selected snippets to an NLE generator, (v)~passes the generated NLE to a predictor that outputs the final answer (see Figs.~\ref{fig:overview} and \ref{fig:variant}). All steps are learned jointly. %and the information flows all the way from the input to the prediction. 
\ours{}~does not require direct supervision on the ER and snippet selections, which are modeled by two series of latent variables and variational learning (\Cref{sec:ours}). Supervision comes from
the final answers and NLEs. %Thus, none of the five steps in \ours{} creates a strong bottleneck.

\ours{} is illustrated in Fig.~\ref{fig:overview}. In Fig.~\ref{fig:overview}b, a subset of super-pixels of an input image form the selected ER for the question-answering instance. % (the instance is taken from the Commonsense Question Answering dataset \cite{DBLP:conf/naacl/TalmorHLB19}). 
To answer that ``\texttt{Person2} is guarding \texttt{person3}'' and explain the answer, the model needs to identify that \texttt{person2} holds a weapon and have the knowledge that weapons are used to protect.

In our experiments spanning natural language (NL) and vision-language (VL) domains, we find that \ours{}~significantly improves the quality of both ERs and NLEs, while bridging the gap between task performance and explainability. %To be specific, we significantly outperform the previous SOTA in NLE generation across five tasks, spanning both natural language (NL) and vision-language (VL) domains. Furthermore, ERs from \ours{}~are also of better quality compared to the ones from previous SOTA models with (only) extractive rationales. 
We also show, via perturbation analysis, that the explanations from \ours{}~exhibit necessary conditions of faithfulness. % \cite{DBLP:journals/corr/abs-2010-12762, DBLP:conf/acl/DeYoungJRLXSW20}, being strongly associated with the predicted labels
%(\Cref{sec:faith}).
Finally, \ours{}~allows the selection of relevant knowledge snippets even without supervision from the NLEs. As these snippets can act as NLEs, we provide a zero-shot model with NLEs (\ours{}-ZS), which proves to be competitive with its supervised version.

The contributions of this work are summarized as follows:
\vspace{-1.2ex}
\begin{itemize}[leftmargin=*, itemsep=0.05pt]
    \vspace{-1ex}
    \item We propose a novel self-rationalizing framework that incorporates background knowledge and provides two complementary types of explanations: ERs and NLEs.
    
    \vspace{-1ex}
    \item \ours{}~consistently outperforms previous best models that produce at least one type of explanation and performs on par with the SOTA models that do not provide any explanation, thus bridging the gap between explainability and task performance.
    
    \vspace{-1ex}
    \item \ours{}~largely outperforms the previous SOTA in NLE and ER quality. %We show that the joint modeling of predictive tasks and explanations can lead to a better predictive performance with new SOTA in two datasets: ComVE~\cite{wang-etal-2019-make} and e-SNLI-VE \cite{xie2019visual, DBLP:journals/corr/abs-2105-03761}.
    
    \vspace{-1ex}
    \item \ours{}~passes necessary faithfulness tests. % by observing a high degree of association between outputs and both types of explanations. 
    
    \vspace{-1ex}
    \item \ours{}~allows for a zero-shot setting in terms of~NLEs (\ours{}-ZS), which sometimes outperforms models~trai\-ned with a full training set of NLEs. 
\end{itemize}

%% file: files/method.tex
% \section{\ours{}:~Rationale-Inspired Explanations with Commonsense}
\section{\ours{}} %:~Self-Rationalization with Knowledge Grounding}
\label{sec:ours}

\begin{figure*}[t!]
    \centering
    \includegraphics[trim= 60 315 150 330,clip, width=0.9\textwidth]{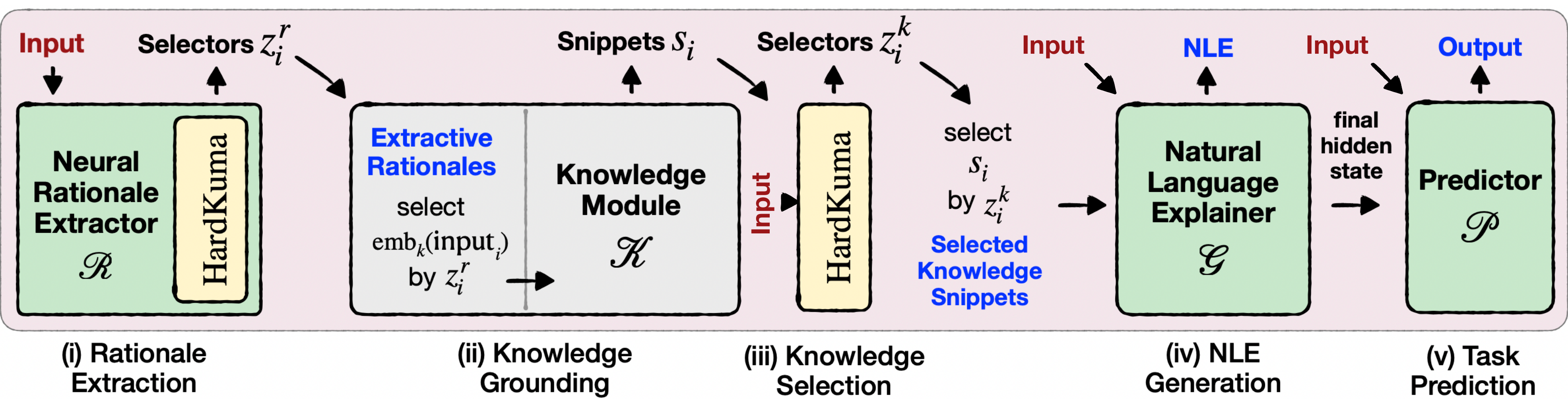}
    \vspace{-0.9ex}
    \caption{\small 
    % \ours{}~extracts rationales from the input as important features for model prediction. Rationales are used to obtain relevant commonsense knowledge. Both rationales and obtained knowledge directly influence the NLE and predicted output. In end-to-end training, $\mathcal{M}_\mathcal{T}$ and $\mathcal{G}$ are updated via backpropagation while $\mathcal{K}$ remains fixed.
    %Joint modeling of rationales, NLEs and outputs in \ours{}~with commonsense.
    \textbf{Architecture of \ours{}.} The knowledge module is frozen, while the rest of the modules are trained jointly with the signals from the NLEs and outputs. Deliverables from \ours{}~are in \textcolor{blue}{blue}.
    }
    % Variants of \ours{}. 
%     % Deliverables
%     Outputs 
%     are shown in \textcolor{blue}{blue}. (a) \textbf{Mod}: $\mathcal{M}_\mathcal{T}$ and $\mathcal{G}$ are trained separately;
%     % each module is trained and tested separately;
%     (b)~%% E2E-\ours{} +
% \textbf{KS-\ours{}}: $\mathcal{M}_\mathcal{T}$ and $\mathcal{G}$ are trained together via backpropagation. The commonsense knowledge %resource 
% $\mathcal{K}$ is fixed in both.} %E2E-\ours{} also selects the most relevant knowledge snippet(s) to guide its generation task.}
    \label{fig:variant}
    \vspace{-0.5em}
\end{figure*}

We aim to build a model that solves a task and explains its predictions via both ERs and NLEs. Furthermore, we aim for our model to benefit from resources of background knowledge, which could be general commonsense or domain-specific. %We ground our model in background knowledge. 
%In this work, we choose commonsense resources for the background knowledge, as that is most appropriate for all our tasks (\Cref{sec:tasks}). However, the architecture of the model allows for domain-specific resources to be used instead or in addition. 
% and bring explanations for the prediction via both extractive rationales and NLEs. More 
To this end, \ours{}~combines these three ingredients in the following way: it extracts rationales from the input, uses them to query an incorporated knowledge module to obtain knowledge snippets, selects the most relevant snippets, generates an NLE, and gives the prediction. We use Fig.~\ref{fig:overview}a as a running example and Fig.~\ref{fig:variant} for an overview of the architecture. % (on an NL task) in this section, while details for both NL and VL tasks are in \Cref{sec:exp}.

% We aim to solve a predictive task given its input  define a neural module selecting the rationale $\mathcal{R}$ that encodes the input (e.g.,~natural language or an image) for a prediction task and extract. Our model brings justifications for its prediction $o$ via both extractive rationales and NLEs. More precisely, the model uses its self-extracted rationales to query a knowledge module whose answers help the NLE generation and the final prediction. 
% motivation is to provide an abstractive justification behind
% the prediction $o$.
% in the form of natural language (NLEs).
%We further use the model-extracted rationales as guiding signals for NLEs. 
% byproduct
% of our framework).
%Our framework combines three ingredients to generate both extractive (rationales) and abstractive (NLEs) explanations. %We introduce various associated concepts to generate such an explanation and then propose our framework to unify them so that the explanation 
% to help
% (non-expert)
% users better understand 
%the 
% machine predictions.

\vspace{-1ex}
\subsection{Extractive Rationales via Binary Latent Variables}
\label{sec:rationales}
\vspace{-0.7ex}
We define a neural module $\mathcal{R}$ that selects an ER from the input. An ER is a minimal sufficient subset of input parts (e.g., tokens for text or super-pixels for images) most responsible for the model's prediction \cite{DBLP:conf/emnlp/LeiBJ16}.
In Fig.~\ref{fig:overview}a, we see an example from the natural language inference task \cite{DBLP:conf/emnlp/BowmanAPM15} (details in \Cref{sec:tasks}), where the ER is %a set of tokens from the premise and hypothesis (
\{``men'', ``people'', ``bicycle race'', ``riding bikes''\}, the most responsible units for the prediction (\textit{entailment}). 

We model the selection of ERs using a series of latent variables ranging from $[0, 1]$ ($z_i^r \in \mathcal{Z}^r$) over the $N$ input units. A unit becomes a part of the ER \emph{iff} its associated variable takes value 1. Following \cite{DBLP:conf/acl/BastingsAT19}, we use the Hard Kumaraswamy distribution (referred to as $\operatorname{HardKuma}$) 
%adapt 
as the reparameterization strategy to learn these latent selectors %directly 
using backpropagation. %Instead of modeling each latent variable with a Bernoulli distribution, we use a Hard Kumaraswamy (referred as $\operatorname{HardKuma}$) distribution \cite{DBLP:conf/acl/BastingsAT19}, which allows binary outcomes and facilitates optimization via backpropagation. 
The parameters of the neural module $\mathcal{R}$ are denoted by $\theta^r$, which estimate the $\operatorname{HardKuma}$ variables for the input units. We also encourage the ERs to be terse, and we control the sparsity using an $L_1$ relaxation defined by the tractable Kumaraswamy~CDF.
%Next, we pass $z_i^r$s to the commonsense module to exposes only the extractive rationale for commonsense expansion.
% The output $o$ is predicted exclusively from the extracted rationales. %---with the goal of making these rationales a likely faithful explanation \cite{DBLP:conf/acl/DeYoungJRLXSW20} (although counter-examples may exist---see the \textit{handshake} concept in \cite{verify}).

% \subsection{Inferring Commonsense Implications about Rationales}
%Extractive rationales are terse and sufficient for machine prediction but may not always form a comprehensive explanation for humans. 
%for the output 
%by a non-expert user. 
%We are therefore also interested in NLEs that are more comprehensive. 
%Commonsense knowledge is crucial for solving and hence for explaining the currently highly-researched NL and VL tasks in the literature \cite{wang-etal-2019-make, DBLP:conf/nips/CamburuRLB18, DBLP:conf/acl/RajaniMXS19, DBLP:journals/corr/abs-2105-03761, DBLP:conf/cvpr/ZellersBFC19}.\footnote{Our model is able to incorporate any type of knowledge module, including more specialized ones for, e.g., legal or medical domains.} %, and explanation.

\vspace{-1ex}
\subsection{Knowledge about an Extractive Rationale}
%Tasks that benefit from complex explanations often require background knowledge to solve.
%them as well. 
We hypothesize that inferred knowledge about the ERs are the most important bits of information for the predictions and, implicitly, for the NLEs. %Here, we use commonsense knowledge 
% \footnote{Our model is able to incorporate any type of knowledge besides commonsense, e.g., for legal or medical domains.} 
%as the background knowledge inferred from the obtained ERs.
% $z_i^r$s to the commonsense module to expose only the extractive rationale for commonsense expansion. 
For example, in Fig.~\ref{fig:overview}a, we obtain relevant knowledge snippets (\textit{bicycle race requires bikes} and \textit{men are people}) for the ER (``bicycle race'', ``men'', ``people''), which influence both the prediction and the NLE.

We use a knowledge module $\mathcal{K}$, which supports input from an appropriate modality (e.g., text or image) for querying.
%as appropriate. 
We query $\mathcal{K}$ with each contiguous element of the ER (e.g., ``bicycle race'') to obtain a large pool of associated knowledge snippets $\mathcal{S}$. We take advantage of recent developments in generative models capable of providing background knowledge about a given entity for the ease of end-to-end training, such as \texttt{COMET} \cite{DBLP:conf/acl/BosselutRSMCC19} for NL inputs and \texttt{VisualCOMET} \cite{DBLP:conf/eccv/ParkBMFC20} for image inputs. 
%For example, given the entity ``car'', \texttt{COMET} would output commonsense knowledge snippets such as ``car is made up of car engine, front wheel, etc.''.
%For each query, $\mathcal{K}$ provides knowledge snippets according to a fixed number of (depending on the knowledge resource) knowledge relations. We denote the parameters of $\mathcal{K}$ as $\vartheta^k$.
% , which are kept frozen during \ours{}'s training. 
% Thus, the commonsense knowledge resource in \ours{}~is a parametric model (a fine-tuned language model on commonsense knowledge bases \cite{DBLP:conf/aaai/SpeerCH17, DBLP:conf/aaai/SapBABLRRSC19}),
%hence do not 
The generative knowledge module does not
suffer from the no-hit issue that is typically encountered in retrieval settings.
However, \ours{}~is flexible to accommodate a retrieval-based knowledge source when equipped with a differential search (see \Cref{sec:retrieval}).
% that is differential.  
% Moreover, the parametric form of the knowledge module  is easily integrable in our end-to-end framework.
To facilitate end-to-end training, we use soft representations of the elements of the ER---which are encoded using the embedding layer of $\mathcal{K}$ and subsequently selected by $z_i^r$ (when 1)
for queries to $\mathcal{K}$. Finally, we denote the parameters of $\mathcal{K}$ as $\theta^k$.

% (i.e. input embeddings  using the embedding layer of $\mathcal{K}$) of the elements in the ER to construct queries for $\mathcal{K}$. 
% % For that, we embed the input using the embedding layer of $\mathcal{K}$ as $\operatorname{emb}_k(\text{input})$ and use 
% We use $z_i^r$ as an indicator to select (when $z_i^r = 1$) the equivalent soft representation for
% ER from input to pass to $\mathcal{K}$.

\subsection{Knowledge Selection}
While the knowledge module generates several knowledge snippets ($\mathcal{S}$), not all of them are relevant for the prediction. Hence, we introduce a knowledge selection step. % before prediction and NLE generation. % can potentially improve the quality of the NLEs. 
Furthermore, the selected knowledge snippets can appear as supporting evidence in addition to the generated NLE---an advantage of \ours{} over models that only generate NLEs.

% The ground-truth selection of knowledge is usually not available; hence,
We model the selection step via another set of latent selectors $z_i^k \in \mathcal{Z}^k$, which take a value from the interval $[0, 1]$ and are 
realized by a $\operatorname{HardKuma}$ distribution (similarly to \Cref{sec:rationales}). More than one knowledge snippet may be relevant, however, we want the knowledge selection to be sparse. %, such that \ours{}~does not default to using all knowledge snippets. 
Hence, we use $L_1$ regularization to control the sparsity of the
selected knowledge. The parameters predicting the latent selectors~$z_{i}^k$ are denoted as $\theta^{ks}$.

To facilitate end-to-end training, we do not decode knowledge snippets into natural language. Instead, we retain the final hidden representations of each snippet from the knowledge module as $s_i \in S$. 
Using $z_i^k$ as an indicator of selection, we obtain the vectors of selected knowledge snippets and 
% $s_i \odot z_i^k$
concatenate them as input to the NLE generator. We also concatenate the representation of the input for the selector to be able to select the most relevant snippets given the input. % (see Fig.~\ref{fig:variant}).
At inference time, we decode the selected knowledge snippets into language, which could be used 
% commonsense and provide them 
as additional supporting evidence along with the NLE. We call this variant \ours{}+. Human evaluation shows that this additional evidence leads to higher quality explanations (\Cref{sec:quality}).

% \begin{figure}[t!]
%     \centering
%     %\includegraphics[width=0.50\textwidth]{images/overview.png}
%     \includegraphics[trim=300 360 860 330,clip, width=\linewidth]{images/tacl_ks.pdf}
%     \caption{\small \textbf{Training:} Soft knowledge selection realized by $\operatorname{HardKuma}$ variables. \textbf{Inference:} Deterministic sampling for high-value selectors.
%     The selected knowledge is decoded into natural language  %using nucleus sampling for supporting evidence in 
%     in \ours{}+. 
%     }
%     \label{fig:ks}
%     \vspace{-0.8em}
% \end{figure}

\vspace{-0.5em}
\subsection{NLE Generation and Task Prediction}
\label{sec:generation}
%The soft representations of the selected knowledge snippets are given to an NLE generator. 
We use a natural language decoder $\mathcal{G}$, which concatenates the soft representations of the knowledge snippets and of the instance input at the input layer
% concatenated at the embedding layer 
% \cite{DBLP:conf/iclr/BhagavatulaBMSH20} 
and %decodes them to 
generates an NLE. After $\mathcal{G}$, we add a predictor module $\mathcal{P}$, a linear layer with softmax, which takes the final hidden representation of the NLE and the representation of the instance input, and projects them to the output space for the task prediction. % (see Fig.~\ref{fig:variant}). 
The prediction is thus directly conditioned on the NLE and the input, and, implicitly, on the ER and selected snippets. We denote the parameters of $\mathcal{G}$ and $\mathcal{P}$ as $\theta^g$ and $\theta^p$, respectively. We use direct supervision from the ground-truth NLEs and task outputs.

\vspace{-0.5ex}
\subsection{Training}

The parameters for $\mathcal{R}$, $\mathcal{G}$, $\mathcal{P}$, and the knowledge selector can be jointly trained end-to-end with backpropagation by summing up the negative log-likelihoods for the predictions and NLEs.
%Note that, since $\mathcal{G}$ produces the output, the $\mathcal{L}^r$ becomes a function of $\mathcal{G}$. 
We found that updating parameters for the knowledge resource $\mathcal{K}$ led to a minimal improvement; hence, $\mathcal{K}$ is fixed for computational ease. 

However, due to the presence of $z_i^r$s in $\mathcal{R}$, we instead have to optimize a lower bound $\mathcal{E}$ of the original log-likelihood. We follow \citet{DBLP:conf/acl/BastingsAT19} and optimize $\min _{\theta^r, \theta^g, \theta^{ks}, \theta^p} \mathcal{L}_1$ with
\vspace{-0.5ex}
\begin{equation}
\label{equ:r}
\begin{split}
% F = \{F_{x} \in  F_{c} &: (|S| > |C|) \\
%  &\quad \cap (\text{minPixels}  < |S| < \text{maxPixels}) \\
%  &\quad \cap (|S_{\text{conected}}| > |S| - \epsilon) \}
\mathcal{L}_1 = -\mathcal{E}(\theta^r, \theta^k, \theta^{ks}, \theta^g, \theta^p) \\ + \lambda_{0}^r \sum\nolimits_{i=1}^{N} z_{i}^r +
\lambda_{1}^r \sum\nolimits_{i=1}^{N-1}\left|z_{i}^r-z_{i+1}^r\right|,
\end{split}
\end{equation}
% \vspace{-1ex}
% \[
% \min _{\vartheta^r, \theta^r}-\mathcal{E}(\vartheta^r, \theta^r)+\lambda_{0}^r \sum\nolimits_{i=1}^{N} z_{i}^r+\lambda_{1}^r \sum\nolimits_{i=1}^{N-1}\left|z_{i}^r-z_{i+1}^r\right|,
% \]
where the second term is the $L_1$ penalty, the third term is a fused Lasso to control the total number of transitions for compactness \cite{DBLP:conf/emnlp/LeiBJ16}, and $\lambda_{0}^r$ and $\lambda_{1}^r$ are hyperparameters. Similarly, we have another lower bound for the $z_i^k$ variables in the knowledge selection step, for which we optimize $min _{\theta^{ks}, \theta^g, \theta^p} \mathcal{L}_2$ with 
\vspace{-0.5ex}
\begin{equation}
\label{equ:ks}
\mathcal{L}_2 = -\mathcal{E}(\theta^{ks}, \theta^g, \theta^p)+ \lambda_{0}^k \sum\nolimits_{i=1}^{M} z_{i}^k, 
\end{equation}
% \vspace{-1ex}
where the second term denotes 
%the 
$L_1$ regularization for sparse knowledge selection. Finally, we combine the lower bounds as $\alpha \times \mathcal{L}_1 +  (1-\alpha) \times \mathcal{L}_2$, where $\alpha \in [0,1]$ is a hyperparameter. We estimate the gradient of $\mathcal{E}$ via Monte-Carlo sampling from the reparameterized $\operatorname{HardKuma}$ variables \cite{DBLP:journals/corr/KingmaW13}. All hyperparameters are chosen based on a greedy search over the task prediction accuracy (more in \Cref{sec:ap_exp}). 

%% file: files/experiments.tex
\section{Experiments}
\label{sec:exp}

\paragraph{Tasks.} We experiment with three tasks of natural language and two tasks of vision-language understanding as described in \Cref{tab:tasks}. More task details are in \Cref{sec:ap_data}.
\label{sec:tasks}

% \vspace{-1.5ex}
\begin{table}[h!]
\caption{\textbf{Our tasks: three NL and two VL.} %Original tasks were introduced by (in order) \citet{wang-etal-2019-make}, \citet{DBLP:conf/emnlp/BowmanAPM15}, \citet{DBLP:conf/naacl/TalmorHLB19}, \citet{xie2019visual}, \citet{DBLP:conf/cvpr/ZellersBFC19}. We use datasets that also provides gold NLEs for these tasks.
}
\label{tab:tasks}
\vskip 0.08in
\small
\centering
\resizebox{\linewidth}{!}{%
\begin{NiceTabular}{@{}lcc@{}}
\toprule
\bf Task\!\!\!                                                                      & \bf \!\!\!\!\!\!Dataset\!\!                                                                             & \bf \!Summary\!                                                                                       \\
\midrule
\begin{tabular}[c]{@{}l@{}}\bf Commonsense \\\bf  Validation\end{tabular}         & \begin{tabular}[c]{@{}c@{}}ComVE\\ \cite{wang-etal-2019-make}\end{tabular}          & \begin{tabular}[c]{@{}c@{}}Choosing input sentence\\ that defies commonsense\end{tabular}      \\[0.3cm]
\begin{tabular}[c]{@{}l@{}}\bf Natural Language\\\bf  Inference\end{tabular}      & \begin{tabular}[c]{@{}c@{}}e-SNLI\\ \cite{DBLP:conf/nips/CamburuRLB18}\end{tabular} & \begin{tabular}[c]{@{}c@{}}Textual entailment between\\ premise and hypothesis\end{tabular}    \\[0.3cm]
\begin{tabular}[c]{@{}l@{}}\bf Commonsense \\\bf Question Answering\end{tabular} & \begin{tabular}[c]{@{}c@{}}COSe\\ \cite{DBLP:conf/acl/RajaniMXS19}\end{tabular}     & \begin{tabular}[c]{@{}c@{}}Answering multi-choice\\ commonsense questions\end{tabular}         \\[0.3cm]
\begin{tabular}[c]{@{}l@{}}\bf Visual\\\bf Entailment\end{tabular}               & \begin{tabular}[c]{@{}c@{}}e-SNLI-VE\\ \cite{kayser21}\end{tabular}                 & \begin{tabular}[c]{@{}c@{}}Entailment between image\\ premise and text hypothesis\end{tabular} \\[0.3cm]
\begin{tabular}[c]{@{}l@{}}\bf Visual Commonsense\\\bf Reasoning\end{tabular}    & \begin{tabular}[c]{@{}c@{}}VCR\\ \cite{DBLP:conf/cvpr/ZellersBFC19}\end{tabular}    & \begin{tabular}[c]{@{}c@{}}Commonsense reasoning in\\ visual question-answering
% \footnote{Treated as a multi-choice classification.}
\end{tabular} \\
\bottomrule
\end{NiceTabular}%
}
\end{table}

\begin{table*}[t!]
% \footnotesize
\caption{\small \textbf{Task performance (Acc.) and NLE quality for the (a) NL and (b) VL tasks.} NLE Automatic metrics: METEOR (MET.), BERTScore (BRTSc.), BLEURT (BLRT.), and NLE human evaluation metrics: e-ViL score, Yes/No \%s. \textbf{Bold} indicates the best numbers 
with statistical significance
% differences
($p < 0.001$). \underline{Underline} indicates best task performance from a model with (any type of) explanations.
% \ours{}+ adds the selected knowledge with the generated NLEs, hence is only used in the human evaluations.
}
\vskip 0.08in
\begin{subtable}[t]{\textwidth}
% \centering
\resizebox{\textwidth}{!}{%
\centering
\begin{NiceTabular}{lccccccccccccccccccccc}
\toprule
& \multicolumn{7}{c}{\bf ComVE} & \multicolumn{7}{c}{\bf e-SNLI} & \multicolumn{7}{c}{\bf COSe}\\
\cmidrule(l{2pt}r{2pt}){2-8} \cmidrule(l{2pt}r{2pt}){9-15} \cmidrule(l{2pt}r{2pt}){16-22}
\bf Model  & \bf \!\!\!Acc.\!\!\! & \bf \!\!\!MET.\!\!\! & \bf \!\!BRTSc.\!\! & \bf \!\!\!BLRT.\!\!\! & \bf e-ViL\!\!\! & \bf Yes & \bf No & \bf \!\!\!Acc.\!\!\! & \bf \!\!\!MET.\!\!\! & \bf \!\!BRTSc.\!\! & \bf \!\!\!BLRT.\!\!\! & \bf e-ViL\!\!\! & \bf Yes & \bf No & \bf \!\!\!Acc.\!\!\! & \bf \!\!\!MET.\!\!\! & \bf \!\!BRTSc.\!\! & \bf \!\!\!BLRT.\!\!\! & \bf e-ViL\!\!\! & \bf Yes & \bf No\\
\midrule
Gold & -- & -- & -- & -- & 91.6 & 79.3 & 1.1 & -- & -- & -- & -- & 98.1 & 94.1 & 2.7 & -- & -- & -- & -- & 84.8 & 74.5 & 1.8 \\
Task SOTA & 97.0 & -- & -- & -- & -- & -- & -- & \bf 93.1 & -- & -- & -- & -- & -- & -- & \bf 83.7 & -- & -- & -- & -- & -- & -- \\\midrule
% Neg-Heu & -- & 1.2 & 78.2 & 21.4 & 77.5 & 47.7 & 1.4 & -- & -- & -- & -- & -- & -- & -- & -- & -- & -- & -- & -- & -- & -- \\
NILE 
% \cite{DBLP:conf/acl/KumarT20} 
& -- & -- & -- & -- & -- & -- & -- & 91.9 & 11.3 & 75.3 & 41.2 & 84.3 & 80.1 & 9.4 & -- & -- & -- & -- & -- & -- & -- \\
CAGE 
% \cite{DBLP:conf/acl/RajaniMXS19}  
& -- & -- & -- & -- & -- & -- & -- & -- & -- & -- & -- & -- & -- & -- & 72.1 & 1.3 & 43.1 & 16.9 & 59.5 & 35.4 & 16.7 \\
WT5 
% \cite{DBLP:journals/corr/abs-2004-14546}  
& 96.1 & 3.4 & 86.4 & 27.0 & 67.7 & 46.2 & 11.0  & 92.1 & 12.3 & 75.3 & 42.3 & 85.3 & 82.7 & 12.8 & 81.0 & 2.2 & 52.0 & 22.4 & 73.0 & 53.9 & 10.5 \\
\midrule
\rowcolor{Gray}
%\rowcolor[gray]{.9}
% Mod-\ours{}
%   &  7.2 & 88.6 & 30.1 & 51.2 &  10.5 & 14.4 &  78.9 &  47.2 &  87.8 & 6.6 & 2.2 & 52.4 & 26.4 & 62.6 &  5.7 \\
\rowcolor{Gray}\ours{}-ZS  & 96.7 &  7.7 &  72.4 &  24.2 & 65.8 &  56.5 &  16.3 & 92.4 & 11.9 &  63.2 &  40.7 & 88.3 & 85.8 &  5.5 & 83.1 &  2.6 & 38.1 &  17.1 & 83.4 & 73.2 &  5.6 \\\midrule
\rowcolor{Gray}\ours{}  & \bf \underline{97.2} &  \bf 14.1 &  \bf 91.9 &  \bf 33.7 & 87.3 &  \bf 72.6 &  2.8 & \underline{92.9} & \bf 19.6 & \bf 86.8 &  \bf 51.3 & \bf 94.9 & 93.9 &  3.6 &  \underline{83.6} & \bf 7.2 & \bf 60.3 &  \bf 30.5 & 87.4 & \bf 74.3 &  2.1 \\
\rowcolor{Gray}\ours{}+  & \bf \underline{97.2} &  -- &  -- &  -- &  \bf 88.4 & \bf 72.6 &  \bf 1.2 & \underline{92.9} & -- &  -- &  -- &  \bf 95.6 & \bf 94.3 &  \bf 2.7 & \underline{83.5} & -- &  -- &  -- & \bf 87.9 & \bf 74.7 & \bf 1.8 \\
\rowcolor{Gray}\ours{}-RB 
% \cite{DBLP:conf/nips/LewisPPPKGKLYR020} 
& 96.4 & 3.1 & 89.5 & 26.1 & 62.2 & 43.3 & 15.1 & 92.7 & 13.2 & 77.4 & 45.3 & 87.6 & 81.2 & 13.5 & 82.2 & 3.7 & 55.5 & 23.8 & 79.3 & 63.2 & 9.6 \\
\rowcolor{Gray}w/o KN-Sel  & 97.1 &  11.3 &  90.2 & 33.6 & 84.4 &  65.3 &  5.1 & 92.8 & 17.9 &  83.4 & 51.2 & 92.8 & 91.7 &  5.8 & 83.2 & 6.4 & 58.4 &  27.9 & 85.0 & 70.2 &  2.5 \\
\rowcolor{Gray}w/o ER
& 96.5 & 5.2 & 86.1 & 28.1 & 67.2 & 43.4 & 7.6 & 92.3 & 13.1 & 77.7 & 43.5 & 83.4 &  83.2 & 15.1 & 81.4 & 2.9 & 52.8 & 23.8 & 66.7 & 45.2 & 14.9 \\
\rowcolor{Gray}w/o KN \& ER
% \cite{DBLP:conf/acl/LewisLGGMLSZ20} 
& 96.0 & 4.3 & 85.2 & 26.3 & 66.6  & 41.3 &7.6 & 92.2 & 12.4 & 76.4 & 41.9 & 82.9 &  81.2 & 15.7 & 80.8 & 2.5 & 51.6 & 22.4 & 65.9 & 44.1 & 15.9 \\
\bottomrule
% \midrule
\end{NiceTabular}%
}
% \vspace{-1ex}
\caption{}
\label{tab:nl}
% \caption{\small \textbf{Task performance (Acc.) and NLE quality for the NL tasks.} Automatic metrics: METEOR (MET.), BERTScore (BRTSc.), BLEURT (BLRT.), and human evaluation metrics: e-ViL score, Yes/No \%s. \textbf{Bold} indicates best numbers
% % significant differences
% ($p < 0.05$). \underline{Underline} indicates SOTA task performance from a model with explanations.
% }
% \label{tab:nl}
\end{subtable}%
\\
\smallskip
\begin{subtable}[t]{\textwidth}
\centering
% \centering
\resizebox{0.78\textwidth}{!}{%
\centering
\begin{tabular}{lcccccccccccccc}
\toprule
& \multicolumn{7}{c}{\bf e-SNLI-VE} & \multicolumn{7}{c}{\bf VCR}\\
\cmidrule(l{2pt}r{2pt}){2-8} \cmidrule(l{2pt}r{2pt}){9-15}
\bf Model  & \bf Acc. & \bf MET. & \bf BRTSc. & \bf BLRT. & \bf e-ViL & \bf Yes & \bf No & \bf Acc. & \bf MET. & \bf BRTSc. & \bf BLRT. & \bf e-ViL & \bf Yes & \bf No\\
\midrule
% & \bf System $\rightarrow$  & \bf Gold & \bf ROUGE & \bf BERTSc. & \bf METEOR & \bf SPICE  & \bf CIDEr & \bf BLEURT & \bf Yes & \bf W-Yes & \bf W-No & \bf No\\
Gold & -- & -- & -- & -- & 90.6 & 79.3 & 1.1 & -- & -- & -- & -- & 95.8 & 94.1 & 2.7\\
Task SOTA & 79.5 & -- & -- & -- & -- & -- & -- & \bf 81.6 & -- & -- & -- & -- & -- & --  \\\midrule
PJ-X 
% \cite{DBLP:conf/cvpr/ParkHARSDR18}
& 69.2 & 14.7 & 79.1 & 35.6 & 70.1 & 55.2 & 14.5 & 39.0 & 16.4 & 78.4 & 43.5 & 73.9 & 58.2 & 10.5 \\
FME
% \cite{DBLP:journals/corr/abs-1809-02805}
& 73.7 & 15.6 & 79.7 & 34.5 & 71.9 & 56.7 & 13.2 & 48.9 & 17.3 & 79.4 & 47.8 & 73.0 &  56.2 & 11.1 \\
RVT
% \cite{DBLP:conf/emnlp/MarasovicBPBSC20}
& 72.0 & 18.8 & 81.1 & 35.3 & 72.2 & 55.4 & 12.8 & 59.0 & 11.2 & 78.9 & 44.2 & 73.2 & 57.4 & 11.5 \\
e-UG
% \cite{DBLP:journals/corr/abs-2004-14546}
& 79.5 & 19.6 & 81.7 & 37.8 & 75.6 & 57.9 & 9.9 & 69.8 & 11.8 & 79.0 & 45.6 & 75.1 & 59.3 & 10.4 \\
\midrule
% \rowcolor{Gray}Mod-\ours{}  &  21.4 & 85.6 & 39.2 & 60.4 & 8.3 & 18.9 & 83.6 & 50.2 & 65.3 & 9.3 \\
\rowcolor{Gray}\ours{}-ZS  &  78.8 & 12.3 &  78.6 & 35.9 &  79.8 & 60.7 & 10.4 & 79.2 & 15.8  & 78.9 & 41.5 & 78.9 & 65.3 &  10.4 \\ \midrule
\rowcolor{Gray}\ours{}  & \bf \underline{80.8} & \bf 22.9 &  \bf 87.7 &  \bf 39.6 & 81.8 & 64.2 & \bf 6.5 & \underline{79.5} & \bf 20.9 &  \bf 86.6 &  \bf 53.1 & 80.9 & \textbf{67.7} &  7.3   \\
\rowcolor{Gray}\ours{}+  & \bf \underline{80.8} &  -- &  -- &  -- &  \bf 82.1 & \bf 65.4 &  6.3 & \underline{79.5} & -- &  -- &  -- & \bf 81.8 & 67.2 &  \bf 6.2 \\
\rowcolor{Gray}\ours{}-RB
% \cite{DBLP:conf/nips/LewisPPPKGKLYR020}
& 78.9 & 20.7 & 83.5 & 38.4 & 78.3 & 59.3 & 10.3 & 78.9 & 14.7 & 81.3 & 47.2 & 78.4 & 62.2 & 11.4 \\
\rowcolor{Gray}w/o KN-Sel  & 79.5 &  22.4 &  86.8 & \bf 39.7 &  79.9 & 62.3 & 7.9 & 78.6 & 19.7 & 85.5 & 51.4 & 79.9 & 67.6 &  8.2 \\
\rowcolor{Gray}w/o ER
& 79.7 & 20.1 & 81.9 & 38.4 & 76.5 & 58.6 & 9.1 & 74.5 & 12.4 & 79.6 & 46.4 & 76.3 & 60.1 & 10.2 \\
\rowcolor{Gray}w/o KN \& ER
% \cite{DBLP:journals/corr/abs-2004-14546}
& 79.4 & 19.5 & 81.7 & 37.7 & 75.5 & 57.9 & 9.8 & 69.8 & 11.9 & 79.0 & 45.8 & 75.1 & 59.4 & 10.5 \\
\bottomrule
\end{tabular}%
}
\caption{}
\label{tab:vl}
% \vspace{-1ex}
% \caption{\small \textbf{Task performance (Acc.) and NLE quality for the VL tasks.} Automatic metrics: METEOR (MET.), BERTScore (BRTSc.), BLEURT (BLRT.), and human evaluation metrics: e-ViL score, Yes/No \%s. \textbf{Bold} denotes significant differences ($p < 0.05$). \ours{}+ adds the selected knowledge with the generated NLEs, hence is only used in the human evaluations.
% }
% \label{tab:vl}
% \vspace{-1em}
\end{subtable}
\vspace{-4ex}
\end{table*}

% -----------------

\paragraph{Implementation Details.} %We discuss the details of the components of \ours{}~when instantiated for solving NL and VL tasks.
%
% \newpara{\ours{}~for NL tasks.}
The components of \ours{}~for the \textbf{NL tasks} are:
% \vspace{-0.8em}
% \begin{itemize}[leftmargin=*, itemsep=0.01em]
%     \setlength\itemsep{0.1em}
\textbf{Rationale extraction:} We use the denoising encoder-decoder \texttt{bart-large} \cite{DBLP:conf/acl/LewisLGGMLSZ20} with a linear layer and softmax at the end to generate the distribution for latent selectors.
\textbf{Knowledge source:} We pre-train a \texttt{bart-large} model as a proxy for \texttt{COMET} (matched with original perplexity, 11.47 vs.~11.14 as from \cite{DBLP:conf/acl/BosselutRSMCC19}) that matches the tokenization scheme used in $\mathcal{R}$.
\textbf{NLE and task output:} We use another \texttt{bart-large} model to generate the NLEs, decoded with top-$p$ sampling ($p=0.95$) \cite{DBLP:conf/iclr/HoltzmanBDFC20}. A linear layer followed by a softmax is used as the task predictor $\mathcal{P}$. % to predict the task label.
% \end{itemize}

% \newpara{Details of \ours{}~for VL tasks.}
\noindent
The components of \ours{} for the \textbf{VL tasks} are:
\textbf{Rationale extraction:} We use a transformer-based VL model, UNITER \cite{DBLP:conf/eccv/ChenLYK0G0020}, which uses self-attention to learn contextualized representations for image-text input pairs. We add two MLPs on top of UNITER, which  are used to generate the distributions for the latent ER selection from the image and text input; \textbf{Knowledge source:} We use \texttt{VisualCOMET} \cite{DBLP:conf/eccv/ParkBMFC20} as an image-based commonsense module, which is fine-tuned on ATOMIC \cite{DBLP:conf/aaai/SapBABLRRSC19}. For text ERs, we follow the same setup as in the NL setup; \textbf{NLE and task output:} %We combine UNITER \cite{DBLP:conf/eccv/ChenLYK0G0020}, an encoder for image-text pairs, and 
We use GPT-2 \cite{radford2019language}, a language decoder, for  NLE generation. %following \citet{kayser21}. 
We adapt GPT-2 to condition on the representations learned by UNITER for VL inputs and use nucleus sampling ($p=0.95$) for decoding the NLEs. A linear layer followed by a softmax is used for task prediction.

\newpara{Baselines.} We consider existing self-explainable models with the SOTA explanations (NLEs or ERs) as baselines. We also compare \ours{}~with models that are SOTA for task performance (all until now are black-box models for our tasks).

\newpara{NL Baselines.\footnote{\label{foot:baseline} We used the implementations from the original works.}}
The current SOTA for NLEs in all three NL tasks was obtained by %We use a general-purpose NLE generation model, 
WT5 \cite{DBLP:journals/corr/abs-2004-14546}, a general-purpose NLE generation model. %as the SOTA for all NL tasks. 
% For e-SNLI and COSe, 
% Further, for each dataset, 
We also compare with works that model NLEs specifically for a dataset: 
% respective
% SOTA models that generate NLEs: 
WT5 for ComVE, NILE \cite{DBLP:conf/acl/KumarT20} for e-SNLI, and CAGE \cite{DBLP:conf/acl/RajaniMXS19} for COSe.

\newpara{VL Baselines.} 
% We consider the suite of models from \citep{kayser21} and report their results. More precisely, we have the following models: % that can work in all VL tasks:  
We compare \ours{}~with: PJ-X \cite{DBLP:conf/cvpr/ParkHARSDR18} and FME  \cite{DBLP:journals/corr/abs-1809-02805}, two self-rationalizing models that provide both NLEs and ERs, and 
% both of which use a model pretrained on VQA tasks; 
RVT \cite{DBLP:conf/emnlp/MarasovicBPBSC20}, a post-hoc explainer that uses external knowledge as \ours.
We also compare with e-UG \cite{kayser21}, the current SOTA in terms of NLE generation on VL tasks.
% which 
% instead of object-level information as RVT,
% uses UNITER to encode VL input and GPT-2 to generate NLEs. 

\newpara{Ablations of \ours{}.}
% Along with the variants of \ours{}~described in Section~\ref{sec:ours}, 
We ablate \ours{}~to investigate the effects of each component: ER selector (w/o ER), knowledge selector (w/o KN-Sel), and both (w/o KN~\&~ER). We also ablate with the NLE generator (\ours{}-ZS), while training just using the final answers as supervision and using the selected knowledge snippets as NLEs. This yields a zero-shot model for NLEs.
\ours{}+ adds the selected knowledge to the NLEs, hence is only used in the human evaluation.
Finally, we also investigate the advantage of the generative knowledge module by replacing it with a retrieval-based knowledge source: ConceptNet \cite{DBLP:conf/aaai/SpeerCH17} and Visual Commonsense Graphs \cite{DBLP:conf/cvpr/ZellersBFC19}. To make the replacement, we use Maximum Inner Product Search as in \cite{DBLP:conf/nips/LewisPPPKGKLYR020}. We call this version \ours{}-RB.

%% file: files/results.tex
\section{Results}
%\subsection{Quality of Explanations} 
\subsection{Evaluating the Quality of the Explanations} 
\label{sec:quality}
% \footnote{Due to lack of space, hyperparameters, additional results on more automatic metrics, human evaluation, and more examples will be given during the code release upon acceptance.}
% and Discussion
% }
We evaluate the quality of the ERs and NLEs for \ours{}~in comparison with the baselines.
% \vfill
% and 
% respect to performance in (1) NLE generation, (2) rationale extraction, and (3) predictive task.
% , and (3) faithfulness.
\vspace{-1.5ex}
\paragraph{Automatic Evaluation of NLEs.}
\vspace{-1ex}
%\paragraph{NLEs.} 
Following \citet{kayser21}, we measure the quality of the NLEs by comparing them with the ground truth when the predicted label is correct. Here, we report METEOR \cite{DBLP:conf/acl/BanerjeeL05}, BERTScore \cite{DBLP:conf/iclr/ZhangKWWA20}, and BLEURT \cite{DBLP:conf/acl/SellamDP20}, which showed the highest correlation with human evaluation \cite{kayser21}. More automatic metrics are reported in \Cref{sec:ap_auto}, \Cref{tab:all}.
% More automatic metrics (following~\cite{DBLP:journals/corr/abs-2105-03761}) are reported in \Cref{sec:ap_auto}.
% The complete table is in the \textcolor{red}{appendix \autoref{?}}.

% BLEU-4 \cite{papineni2002bleu}, ROUGE-L \cite{DBLP:conf/acl/LinO04}, BERTScore \cite{DBLP:conf/iclr/ZhangKWWA20}, METEOR \cite{DBLP:conf/acl/BanerjeeL05}, SPICE \cite{DBLP:conf/eccv/AndersonFJG16}, CIDER \cite{DBLP:conf/cvpr/VedantamZP15}, and BLEURT \cite{DBLP:conf/acl/SellamDP20}
% \begin{itemize}[leftmargin=*, noitemsep]
% \vspace{-0.5em}
For NL tasks, 
%we observe 
\ours{}~achieves the best values 
% increasingly perform trend toward \ours{}~
on all three automatic metrics (see \autoref{tab:nl}). We see sharp jumps (e.g., ranging from 4.8 to 11 points in METEOR)
% 17-29\%)
between \ours{}~and models that do not use knowledge grounding, such as \ours{} w/o KN~\&~ER 
and WT5. 
% This confirms that commonsense is effective in generating high-quality NLEs.
% Outperforming 
% fine-tuned versions of pretrained language models (e.g.~T5 vs.~\ours{})
This confirms that background knowledge is a useful %necessary
component for better NLEs.
The gains for \ours{}~over \ours{}~w/o KN-Sel.~show that knowledge selection provides a regularizing effect. % aligning extractive rationales and outputs to the NLEs.
% to  that jointly producing rationales with NLEs has a clear advantage.
    
Similarly, \ours{}~outperforms the previous
% state-of-the-art
SOTA models for VL tasks (see \autoref{tab:vl}). 
% we observe a similar trend in 
    %to NL 
% tasks. 
In particular, \ours{}~outperforms RVT, a competitive model providing post-hoc NLEs also using the same commonsense resource as \ours, which possibly indicates that joint training for predictions and NLEs is superior over a post-hoc explainability approach.  
% \end{itemize}

\begin{figure*}[t!]
    \centering
    \includegraphics[trim= 45 580 220 55,clip, width=0.95\textwidth]{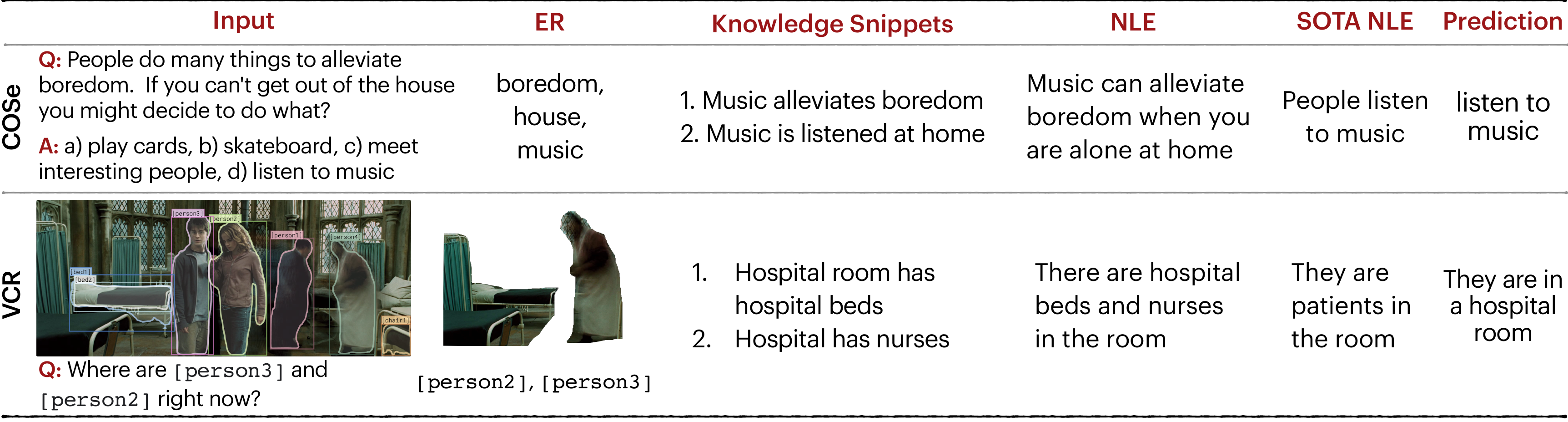}
    \vspace{-0.8ex}
    \caption{\small \textbf{Examples of NLEs and ERs} generated from \ours{}~along with selected knowledge snippets
    % five tasks, 
    vs.~those from the previous SOTA for the correct predictions for COSe and VCR. Error analysis (\Cref{fig:error}) and more examples (\Cref{fig:example2}) are in \Cref{sec:ap_human}. %NLEs from \ours{}~\colorbox{yellow}{follow} ERs and grounded in background knowledge.
    % selected by \ours{}. 
    % Generations from the best baseline are included for direct comparison. Both SOTA and \ours{}~produce correct prediction, as shown here.
    } 
    \label{fig:example}
    \vspace{-0.5em}
\end{figure*}

\begin{table}[t!]
    \caption{\small \textbf{ER quality.} Comparison of previous SOTA models \cite{DBLP:conf/acl/DeYoungJRLXSW20} for rationale extraction vs.~\ours{}~for ER quality. %on e-SNLI and COSe, 
    % using the ERASER evaluation framework \cite{DBLP:conf/acl/DeYoungJRLXSW20}. 
    % Mod-\ours{} is performing
    % potentially 
    % on par with the standalone model that only performs rationale extraction and the predictive task.
    Best numbers are in \textbf{bold}.}
    \label{tab:rationale}
    \vskip 0.08in
    \centering
    \resizebox{0.85\linewidth}{!}{%
    \begin{tabular}{lcccccc}
    \toprule
     &  \multicolumn{3}{c}{\bf e-SNLI} & \multicolumn{3}{c}{\bf COSe}\\
    \cmidrule(l{2pt}r{2pt}){2-4} \cmidrule(l{2pt}r{2pt}){5-7}
    \bf System & \bf Acc. & \bf IOU & \bf Tok. & \bf Acc. & \bf IOU & \bf Tok.\\
    \midrule
    SOTA           &   73.4  &  70.5  &   70.2   & 34.6  &  38.9  &   51.9 \\
    % Base*       &  74.5     &  71.5  &   72.4    & 36.7 & 39.5 & 54.5 \\
    % \rowcolor{Gray}Mod-\ours{} & 74.5 & 71.5  &  72.4  &   36.7    & 39.5 & 54.5 \\
    \rowcolor{Gray}\ours{}      &  \bf 78.4  & \bf 72.9  &  \bf 73.5  & \bf 39.6 & \bf 41.7 & \bf 56.1\\
    \rowcolor{Gray}w/o KN-Sel.  &  77.8 &  72.5 &  73.1 &  38.7 & 40.6 & 55.7\\
    % \rowcolor{Gray}w/o KN & 73.4  &  70.9  &   69.8    & 35.4  & 39.1 & 52.6\\
    \bottomrule
    \end{tabular}%
    }
    \vspace{-3ex}
\end{table}

% \vspace{-1ex}
\paragraph{Automatic Evaluation of ERs.}
% \ours{}~outputs extractive rationales as a byproduct.
To evaluate the quality of ERs, we directly compare them with gold ERs using ERASER \cite{DBLP:conf/acl/DeYoungJRLXSW20}.
% a framework to compare ER vs.~ground-truth. 
ERASER uses accuracy (Acc.) and overlap-based metrics such as F1 at Intersection-Over-Union spans (IOU) and token (Tok.) overlap. % (details in \cite{DBLP:conf/acl/DeYoungJRLXSW20}).
% to measure the overlap.
% between extracted and the ground-truth rationales.
In \autoref{tab:rationale}, we show results for e-SNLI and COSe, the only ones from our list that 
have gold ERs available.
We observe that \ours{}~leads to significantly superior-quality ERs compared to models that do not use NLEs or background knowledge to influence rationale extraction (e.g.,~56. vs.~51.9 F1). Thus, \ours{}~achieves a new SOTA in ERs for both datasets. Possible explanations for this are: (1) additionally optimizing for NLEs constrains \ours{}~to generate more informative ERs, and (2) to obtain better-suited knowledge snippets, \ours{}~must extract high-quality ERs.
% result in better-suited commonsense knowledge for NLEs.

% \vspace{-1.5ex}
\newpara{Human Evaluation of NLEs.}
%\label{sec:user_study}
%\newpara{User study.}
Following \citet{kayser21}, we asked human annotators to 
% designed a user-study to 
measure the quality %effectiveness 
of the generated NLEs. % in explaining the predicted label. % in the context of the input (text for NL tasks and  image-text pair for VL tasks). 
% We make sure that the human annotators are able to solve the predictive task before they evaluate the NLEs.
% thus putting them into the right mindset to evaluate NLEs. We also use this as a quality check for NLE annotation.
% if they could not solve $\mathcal{T}$ correctly in the first place.
For each NLE, we asked: \textit{Given the input, does the explanation justify the answer?} and provide four options: Yes, Weak-Yes, Weak-No, and No. We report the e-ViL score from \cite{kayser21} combining results for each option with a weight of $1$, $\frac{2}{2}$, $\frac{1}{3}$, and $0$ respectively. We only evaluate NLEs for correct predictions and collect 250 random such examples for each model and each dataset. More details are in \Cref{sec:ap_human}.

For NL tasks, \autoref{tab:nl} shows that humans also rated the NLEs from \ours{}~far better than those from the previous SOTA models. Again, \ours{}~without knowledge selection shows large drops, which indicates that the knowledge selection step has positive effects on the quality of the NLEs.
%While Neg-Heu for ComVE gets higher eViL score as compared to other baselines, the NLEs are (as expected) rated far more trivial than those from other models by humans (15\% vs.~3\% for \ours{}). %Even though NLEs from Neg-Heu may appear correct due to their construction, they remain trivial.
%Gains for \ours{}~are significant over previous SOTA models for both ComVE and COSe, which could be due to direct involvement of commonsense in the prediction tasks. On the contrary, we find that explanations for the `neutral' and `contradiction' labels in the e-SNLI dataset often involve reasoning that is beyond the range of commonsense spanned by the input.

For VL tasks, NLEs from previous SOTA models were rated far lower than ground truths, indicating an even bigger need for improvement. 
We observe substantial gains
% for NLEs
for \ours{}, even when compared to competitive models that already use external knowledge, such as RVT \cite{DBLP:conf/emnlp/MarasovicBPBSC20}. %This strengthens our hypothesis that background knowledge about ERs plays an important role in our framework. %rendering better-quality NLEs, and that commonsense knowledge establishes a critical bridge between extractive rationales and NLEs. 
%them.  
% \end{itemize}

% shows we achieve SOTA in rationale extraction.

Often NLEs generated by \ours~are longer than those from the baselines, since they are rich in background knowledge. 
In the human evaluation sample for e-SNLI, we found that 73\% of NLEs from \ours~are longer (at least by a token) compared to NLEs from WT5. However, we find that for \ours{}, length is loosely correlated with the e-ViL score with a Pearson's correlation score of 0.21. This correlation is similar (0.17) for NLEs from WT5. % the previous SOTA. 
We also find similarly low correlations (0.13, 0.24, 0.14, and 0.20) between length and e-ViL score for ComVE, COSe, e-SNLI-VE, and VCR, respectively, which indicates that NLE length did not act as a confounding factor during human evaluation.

% \newpara{Qualitative analysis.}
% \autoref{fig:example} shows sample outputs from \ours{}~for Commonsense QA (NL) and VCR (VL) tasks (more in \Cref{sec:ap_human}).
% % Reflecting our observations from the human evaluation (\Cref{fig:error}), 
% We observe that NLEs generated from \ours{}~are more grounded in knowledge than the NLEs from previous SOTA models (e.g.,~``\textbf{Music} can \textbf{alleviate} boredom when you are \textbf{alone} at home'' in COSe). \ours{}~learns to select a relevant set of knowledge snippets to use 
% %them 
% as supporting evidence for both NLEs and prediction.
% % while grounding NLEs in commonsense. %---an important feature missing in prior works. 
% Moreover, previous SOTA models on generating NLEs
% %lacks in 
% fall short of
% generating comprehensive NLEs (e.g.,~``People listen to music'' for COSe), which could be because they do not condition on ERs (e.g.,~``boredom'').

% \vspace{-1ex}
\newpara{Qualitative Analysis.}
Fig.~\ref{fig:example} shows sample outputs from \ours{}~for COSe and VCR (more in \Cref{sec:ap_human}).
% Reflecting our observations from the human evaluation (\Cref{fig:error}), 
We observe that NLEs from \ours{}~are more grounded in knowledge than those from previous SOTA models.
% (e.g.,~``\textbf{Music} can \textbf{alleviate} boredom when you are \textbf{alone} at home'' in COSe). \ours{}~learns to select a relevant set of knowledge snippets to use 
%them 
% as supporting evidence for both NLEs and prediction.
% while grounding NLEs in commonsense. %---an important feature missing in prior works. 
Moreover, previous SOTA NLEs
%lacks in 
fall short of being comprehensive NLEs (e.g.,~``People listen to music'' for COSe), which could be because they do not condition on ERs (e.g.,~``boredom'').

% \vspace{-1ex}
\subsection{Task Performance}
%performances.}
%Previous work \cite{DBLP:conf/acl/KumarT20} observed that, for SNLI, NLEs can provide an important learning signal to improve the original task performance. To test this hypothesis over a diverse range of tasks and to investigate if commonsense knowledge has a role to play, we compare \ours{}~with SOTA models for each predictive task. 
% So far, 
Until now, the SOTA models in terms of task performance for all five tasks were models that do not offer any explainability \cite{wang-etal-2020-semeval, DBLP:journals/corr/abs-2104-14690, DBLP:conf/iclr/LanCGGSS20, xie2019visual, DBLP:journals/corr/abs-2006-16934}. % (see \autoref{tab:task}). 
Models that attempt to offer explanations (NLEs or ERs) faced a drop in accuracy (see \Cref{tab:nl,tab:vl}). 
%In \autoref{tab:task}, we show that
\ours{}~bridges this important gap by matching SOTA task performance for 4 out of 5 tasks and even achieving a new SOTA for e-SNLI-VE, while providing two types of explanations, both of which are of higher quality than the previous models with SOTA explanations. 
%As evident, 
%Thus, \ours{}~outperforms, in terms of task performance, all previous models that offer explanations. 
%consistently outperforms best performing models that generates some form of explanation along with predictions across all tasks.
%We further achieve a new SOTA in task performance for ComVE and SNLI-VE while comparaing with best performing model that do not provide any explanations.
%Thus, \ours{}~bridges an important gap between explainablity and predictive performance. % by achieving both high task performance and better-quality explanations.
% while we maintain performance in other tasks as compared to current SOTA.
% (BART for NL tasks, and UNITER for VL tasks)
% indicating that NLEs can enhance a model's predictive performance.
% We further achieve a new SOTA in task performance for ComVE and SNLI-VE, while we maintain performance in other tasks as compared to current SOTA.%, strengthening our initial hypothesis. 

\vspace{-1ex}
\subsection{Zero-shot NLEs}
Often, there exists a high overlap between the generated NLEs and the selected knowledge snippets. This is expected, since the NLEs and predictions are conditioned on the selected knowledge.
% in \ours{}.
This raises the question of whether the selected snippets alone could form sufficient NLEs. We argue that, in general, this is not the case, because the information in a background resource may not provide the whole reasoning behind a prediction. This information is only meant to add value but not replace the NLEs. However, in particular cases where the ground-truth NLEs consist mainly of pieces of background knowledge, selected snippets may be sufficient explanations. 
To investigate this for our datasets, we look at \ours{}-ZS, where relevant knowledge was selected \textit{only} using the task prediction loss and concatenated to be used as NLEs. \Cref{tab:nl,tab:vl} show that \ours{}-ZS performs poorly in automatic metrics, which is mostly due to being out of distribution w.r.t.\ the ground-truth explanations. %lack of fluency and compactness. 
However, in human evaluation, we see that even if the NLEs from \ours{}-ZS were not better than the generated NLEs from \ours{}, they were largely better than the NLEs from the previous SOTA models (which were trained with full training sets of NLEs) for 4 out of the 5 tasks. 
% However, the human evaluation shows that while concatenation of commonsense snippets hampers the fluency and they are less preferred than NLEs from \ours{}, annotators preferred these concatenated knowledge snippets more than NLEs from all baselines for across tasks.
These results indicate that: (1) the NLE module in \ours{}~acts as an important conditional generation step that makes NLEs fluent and more comprehensible; and (2) despite being less fluent, concatenated knowledge snippets can act as NLEs in cases where ground-truth NLEs are not present. This shows the potential of \ours{}~for zero-shot natural language rationalization.

% \newpara{Ablation study for \ours{}.}
% To pinpoint the contributing factor behind the improved performance in NLE generation (\Cref{tab:nl,tab:vl}) and in the predictive task (\autoref{tab:task}), we individually drop modules responsible for rationale extraction and commonsense expansion and study relative performance changes. For both NL and VL tasks, we find that an ablative baseline (BART/T5 for NL and RVT for VL) that does not use external commonsense resources consistently lags in NLE quality compared to 
% % all variants of 
% \ours{}. Similarly, in \autoref{tab:task}, we observe performance drops in all prediction tasks (e.g.,~2.7 and 2.5 points in VCR) when we ablate with commonsense and rationale extraction, respectively. Without knowledge selection, the performance of \ours{}~drops both for NLEs and task predictions. 
% Hence,  (1) NLEs (and not just the %external
% commonsense) can enhance a model's predictive performance and (2) knowledge selection acts as a useful inductive bias in effectively utilizing a large pool of knowledge.

\vspace{-1ex}
\subsection{Generative vs.\ Retrieval-based Knowledge Module}
\label{sec:retrieval}
One of the reasons for choosing a generative knowledge module (\texttt{COMET}~and \texttt{VisualCOMET}) is to avoid the no-hit issue of indexed knowledge bases. % or databases. 
For example, when we replaced \texttt{COMET} with ConceptNet \cite{DBLP:conf/aaai/SpeerCH17}, for e-SNLI, we found that 23\% of instances do not retrieve any knowledge snippet. % for any of the rationale-based queries. 
%We also compared \ours{}~with a retrieval-augmented generative (\ours{}-NP) framework trained using Maximum Inner Product Search \cite{DBLP:conf/nips/LewisPPPKGKLYR020} that uses ConceptNet and Visual Commonsense Graphs instead of generative counterparts. 
As expected, \ours{}-RB performed worse than \ours{}~(see \Cref{tab:nl,tab:vl}). %, possibly due to no-hit issues and lack of diversity of the retrieved knowledge.

%% file: files/faithfulness.tex
\section{Evaluating Faithfulness}
\label{sec:faith}

Evaluating the faithfulness of explanations is a challenging open question for both ERs \cite{DBLP:journals/tacl/JacoviG21} and NLEs \cite{DBLP:journals/corr/abs-2010-12762}. We analyze \ours{}~for faithfulness based on existing works. 
 
\begin{figure}[t!]
    \centering
    \includegraphics[width=0.90\linewidth]{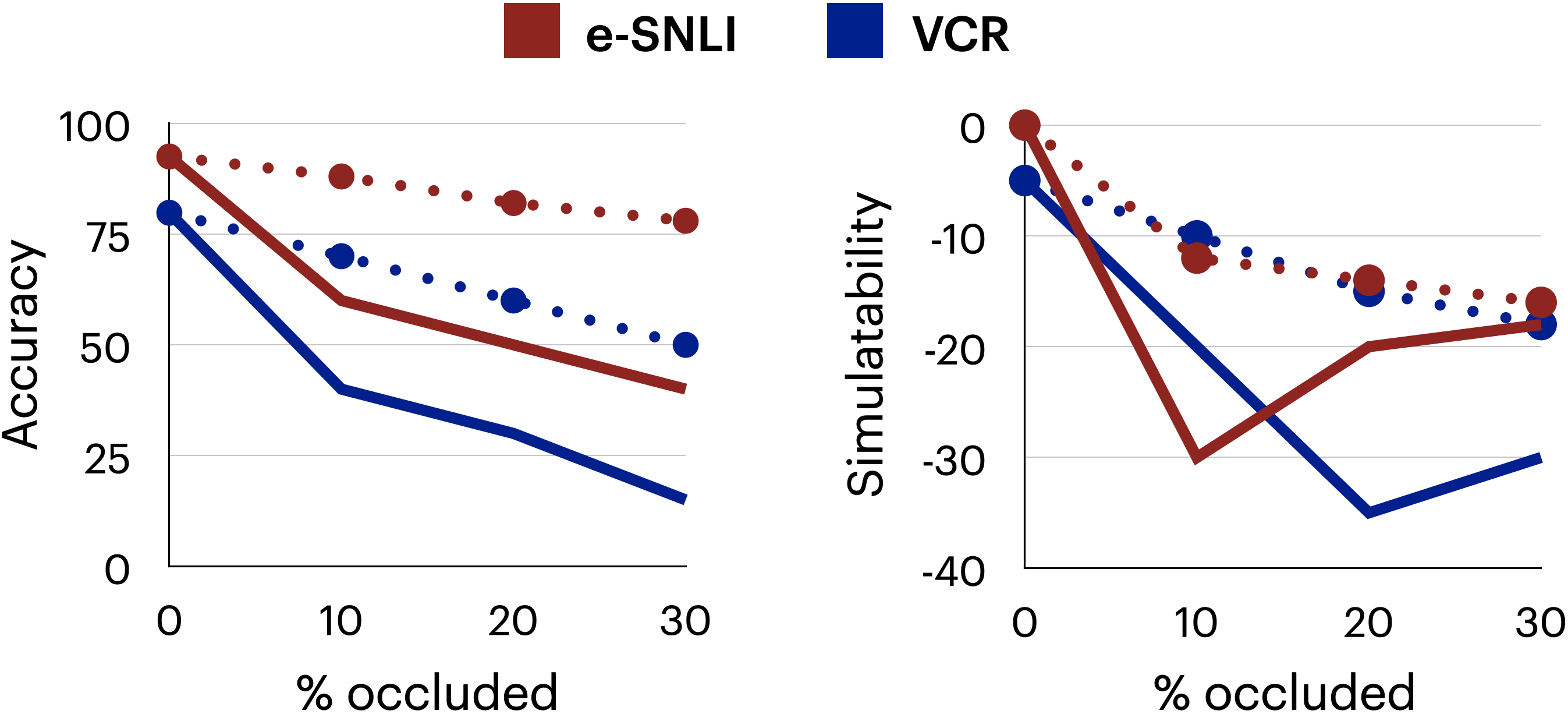}
    \vspace{-0.5ex}
    \caption{\small \textbf{Feature importance agreement.} Left: Solid lines indicate the prediction accuracy when features important for NLEs are occluded. The dotted lines indicate the prediction accuracy when random features are dropped. Right: solid lines indicate the simulatabilities when features important for prediction are occluded. Dotted lines indicates simulatabilities for random occlusions. In both, solid lines should be lower (meaning higher changes) than dotted lines for better label-NLE association.
    % For better label-NLE association, the solid lines both in (c) and (d) should have lower values than the dotted lines.
    % selected by \ours{}. 
    % Generations from the best baseline are included for direct comparison. Both SOTA and \ours{}~produce correct prediction, as shown here.
    } 
    \label{fig:feature_agreement}
    \vspace{-0.5em}
\end{figure}

\subsection{Faithfulness of the NLEs}
Evaluating the faithfulness of NLEs is still in its infancy. To our knowledge, \citet{DBLP:journals/corr/abs-2010-12762} is the only work that provides (two) necessary conditions for NLEs' faithfulness: \textit{feature importance agreement} and \textit{robustness equivalence}. 
%Inspired by \citet{DBLP:conf/nips/HookerEKK19} and following \citet{ DBLP:journals/corr/abs-2010-12762}, we analyse the faithfulness of the NLEs from \ours{}. 
Both conditions perturb the input and measure the change in model behavior in order to establish the extent of label-NLE association. As they mentioned, there are currently no
% universal 
sufficient conditions for faithful NLEs, since there can be different realizations of NLEs that significantly (but differently) contribute to the model's prediction process.

% \vspace{-1ex}
\newpara{Changes in Model Behavior.}
Change in model behavior can be captured by changes in task accuracy and changes in the predictive ability of NLEs. The predictive ability of NLEs over inputs (formally termed as \textit{simulatability} \cite{doshi2017towards, DBLP:conf/emnlp/HaseZXB20}) is defined by the change in task accuracy when the generated NLEs are appended to the input. % (usually at the end). 
To ensure NLEs' faithfulness, changes in accuracy and in NLEs (via simulatability) should be similarly affected by changes in the input. %However, this only indicates a necessary condition for faithful NLEs. % indicating potential for \ours{}'s explanation to be faithful. 
%Sufficient conditions for faithful NLEs are not universally valid for most tasks, since there can be different realizations of NLE that can significantly (but differently) contribute to the model's prediction process \cite{DBLP:journals/corr/abs-2010-12762}. 

% \vspace{-1ex}
\newpara{Feature Importance Agreement.}
This condition uses a gradient-based attribution technique to find the most important features with respect to an output (prediction or NLE). For a predicted class, a gradient attribution is the gradient of the predicted class's logit with respect to an input feature. The attribution score is calculated by performing an operation (here, $L_1$ norm) to turn the gradient into a scalar quantity. For \ours{}, we identify salient input features (tokens or super-pixels) with attribution scores (top-$\{10, 20, 30\}$\%) with respect to the task prediction. We measure the change in simulatability of NLEs when we remove these features %responsible for the task prediction 
from the input. 
Similarly, we measure the change in task accuracy when we remove the features most important for the NLE generation. To ensure faithfulness, both these changes should be significantly higher than the changes that would appear if we were to remove random input features. 
Fig.~\ref{fig:feature_agreement} shows that the removal of salient input features similarly affects both task accuracy and NLEs simulatability when compared to random removal---ensuring that this faithfulness condition is met by \ours{}~on e-SNLI and VCR. Similar trends on the other datasets are in \Cref{sec:ap_faith},~Figure~\ref{fig:feature_agreement_all}.

\begin{figure*}[t!]
    \centering
    \includegraphics[width=0.9\textwidth]{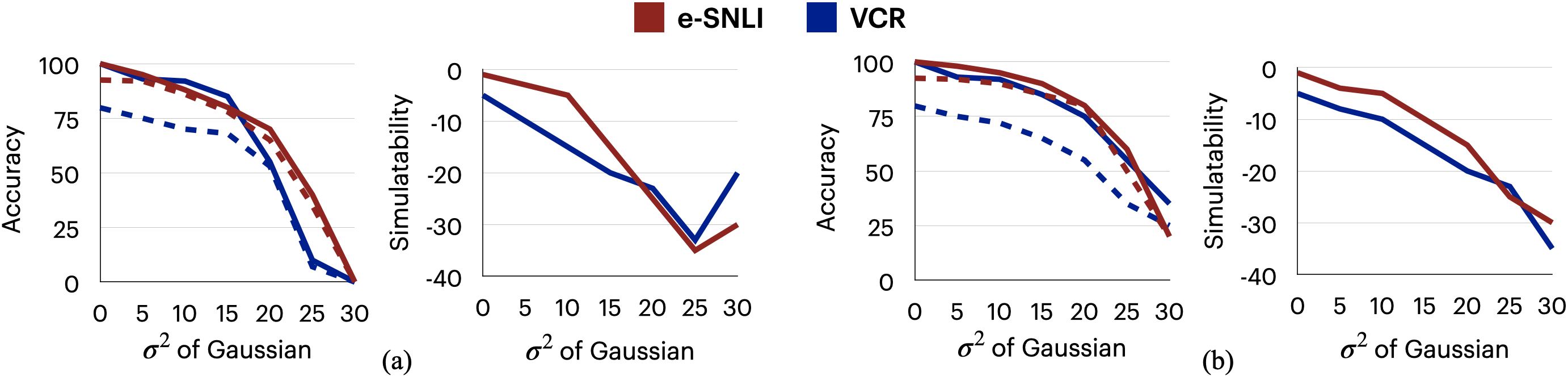}
    \vspace{-1em}
    \caption{\small \textbf{Robustness equivalence} analysis when noise (with various $\sigma^2$) is added to the (a) input and (b) selected knowledge snippets. In each pair, the left chart shows \% of stable (unflipped) labels as the solid line, and accuracy of \ours{}~as the dashed line. The right chart in a pair depicts the simulatability of NLEs. For better label-NLE association, the sharpest drop in simulatability and task accuracy should align with the sharpest drop in \% of stable labels, so that both the labels and the NLEs are stable (or unstable) in the same noise region.
    } 
    \label{fig:faith}
    \vspace{-1em}
\end{figure*}

% In Fig.~\ref{fig:faith}c, we see, for all datasets, that the occlusion of the input parts important for NLE generation makes significant drops in task accuracy. Similarly, in Fig.~\ref{fig:faith}d, the occlusion of the input parts that are important for task prediction contributes to large drops in NLE simulatability across datasets. Thus, our experiments confirm that \ours{}~exhibits a strong association between task outputs and NLEs, passing the faithfulness sanity check as proposed in \cite{DBLP:journals/corr/abs-2010-12762}.

% \vspace{-1ex}
\newpara{Robustness Equivalence.} The second necessary condition involves perturbing the input by adding zero-mean Gaussian noise $\mathcal{N}(0, \sigma^2)$ to the internal representations of its features % and selected knowledge snippets % at inference 
and observing the corresponding changes in task accuracy and NLE simulatability for a range of noise values. We are interested in noise regions where labels and NLEs remain stable (small changes) and noise regions where labels and NLE become unstable (large changes).
To indicate faithfulness of the NLEs, predicted labels and NLEs should remain stable (or unstable) at the same noise region. 
In Fig.~\ref{fig:faith}, we see this condition holds true for \ours{}. For example, for e-SNLI (in Fig.~\ref{fig:faith}(a)), we see that the point of minimum contribution of NLEs to the prediction coincides with the sharpest drop in task accuracy, at $\sigma^2 = 25$. Lower noise than $\sigma^2 = 25$ keeps both labels and NLEs stable, whereas higher noise will make both unstable. Similar trends are observed in other datasets (\Cref{sec:ap_faith}, \Cref{fig:faith-noise-all}).

\subsection{Faithfulness of the ERs and Knowledge Snippets}
For ERs, faithfulness metrics are more studied than NLEs in the literature \cite{DBLP:conf/acl/DeYoungJRLXSW20, DBLP:journals/tacl/JacoviG21}, and both necessary and sufficient conditions for faithfulness exist. %We can measure whether the ERs contain both the necessary and the sufficient signal for the prediction.
\citet{DBLP:conf/acl/DeYoungJRLXSW20} introduced two metrics for measuring faithfulness in ERs: \textit{comprehensiveness} (necessary condition) and \textit{sufficiency}. Comprehensiveness is measured by the change in task accuracy between the case when the full input is used for the prediction by the original model and the case when the ERs (from the original model) are dropped (masked for images) and the model is retrained on these new instances (with dropped ERs).  %before during training. 
A higher difference (maximum 1) would indicate a higher extent of faithfulness. Sufficiency can be calculated as the difference in accuracy between the case when the full input is used for the prediction and the case when \textit{only} the ERs (from original model) are used to retrain the model. A closer to zero value indicates a higher degree of faithfulness. 
For \ours{}, we extend this to the selected knowledge snippets to also analyze their comprehensiveness and sufficiency for the task prediction. %To drop important knowledge snippets, we drop the internal representations of the snippets that were selected by \ours{}'s original training and retrain to measure the change in task accuracy.
\Cref{tab:rationales-faith} confirms solid comprehensiveness (high values) and sufficiency (close to zero) for both ERs and selected snippets. 
%In \Cref{tab:rationales-faith}, high values for comprehensiveness for both ERs and selected snippets indicate their strong association with the predictions. We also observe good sufficiency values (close to zero) for both ERs and selected snippets. 

A baseline for checking faithfulness of ERs and knowledge selection is to check their sufficiency and comprehensiveness with respect to a \textit{random} selection of input tokens as ER and a random selection of knowledge snippets. \Cref{tab:rationales-faith} shows that \ours{} achieves better comprehensive and sufficiency as compared to a random baseline. \ours{} also outperforms all models reported in \citet{DBLP:conf/acl/DeYoungJRLXSW20} in both metrics.
% For e-SNLI, we find random ERs achieve 0.11 for comprehensiveness and 0.54 for sufficiency as compared to 0.45 and 0.08 for ERs from \ours{}, respectively. Similarly, random knowledge snippets score 0.14 (vs.~0.49 for \ours{}) for comprehensiveness and 0.51 (vs.~0.09 for \ours{}) for sufficiency. Higher values for comprehensiveness and lower values for sufficiency are desired, hence \ours{}~achieves significant faithfulness for ERs and selected knowledge snippets. A similar trend is observed in all other datasets. Our results are also better than the models from the ERASER benchmark (DeYoung et al., 2020).

\begin{table}[t!]
\vspace*{-1ex}
\caption{\small \textbf{Comprehensiveness (Comp.) and Sufficiency (Suff.)} metrics for ERs and selected knowledge snippets generated by
    % that indicate the faithfulness of the extractive rationales for
    \ours{} vs.~random ERs and knowledge snippets} 
    \label{tab:rationales-faith}
    \vskip 0.08in
  \resizebox{\linewidth}{!}{%
    \centering
    % \begin{minipage}{0.45\textwidth}
    \small
    \resizebox{\textwidth}{!}{%
    % \begin{tabular}{l@{\hskip -0.1em}c@{\hskip 4pt}c@{\hskip 4pt}c@{\hskip 4pt}c@{\hskip 4pt}c}
    \begin{NiceTabular}{llccccc}
    \toprule
              & & \bf \!\!ComVE\!\! & \bf \!\!e-SNLI\!\!  & \bf \!\!COSe\!\! & \bf \!\!e-SNLI-VE\!\! & \bf \!\!VCR\!\! \\ \midrule
    \textbf{ERs}\\
    Random & Comp.            &  0.12     &  0.11  &   0.10   & 0.13      & 0.14 \\
    \ours{} & Comp.            &  \bf 0.32     &  \bf 0.45  &  \bf 0.24   & \bf 0.28      & \bf 0.33 \\
    Random & Suff.      &  0.44     &  0.31  &   0.54    & 0.51      & 0.39 \\
    \ours{} & Suff.      &  \bf 0.14     &  \bf 0.08  &  \bf 0.05    &  \bf 0.10      & \bf 0.13 \\
    \midrule
    \multicolumn{7}{l}{\textbf{Knowledge Snippets}}\\
    Random & Comp.            &  0.12     &  0.14  &   0.14   & 0.10      & 0.09 \\
    \ours{} & Comp.            &  \bf 0.56     & \bf 0.49  &  \bf 0.36   & \bf 0.27      & \bf 0.35 \\
    Random & Suff.       &  0.41     &  0.51  &   0.43    & 0.51      & 0.37 \\
    \ours{} & Suff.       &  \bf 0.15     & \bf 0.09  & \bf  0.08    & \bf 0.07      & \bf 0.08 \\
    \bottomrule
    \end{NiceTabular}%
    }
    }
\vspace{-1.2em}
\end{table}

%% file: files/related_works.tex
\section{Related Work}
Providing explanations for a model's predictions can be done either post-hoc (via methods that aim to explain already trained and fixed black-box models) or by building self-explainable models (by jointly producing predictions and explanations). Post-hoc explanations \citep{shap, lime} can be useful when one only has access to a high-performance\footnote{High performance on held-out sets does not guarantee that the models do the right thing for the right reasons \citep{rightforwrong}.} but black-box model. However, post-hoc explanatory methods have been shown to have certain downsides \citep{saliency_NEURIPS2018, foolinglimeshap, dangersposthoc, Camburu21WorkshopAAAI, DBLP:journals/corr/abs-2010-12762, verify}. Moreover, self-explanatory models may benefit from the rich information in the explanations provided at training time \citep{betterwithexpl, explainguide, DBLP:journals/corr/abs-2203-05115}.
In this work, we focus on self-explainable models to produce two predominant types of explanations: NLEs and ERs.

% \vspace{-1ex}
\newpara{NLEs.}
A growing number of works in NL and VL focus on designing neural models that produce NLEs for their predictions to make these models accessible to their users \cite{DBLP:conf/eccv/HendricksARDSD16, DBLP:conf/nips/CamburuRLB18, DBLP:conf/cvpr/ParkHARSDR18, kayser21, DBLP:conf/eccv/KimRDCA18, DBLP:conf/acl/LingYDB17, DBLP:conf/emnlp/MarasovicBPBSC20, wang-etal-2019-make, DBLP:conf/acl/RajaniMXS19, DBLP:conf/cvpr/ZellersBFC19}. Recently, \citet{DBLP:journals/corr/abs-2004-14546} achieved SOTA on NLEs for NL tasks by using a 
% transformer encoder-decoder 
%very large 
pre-trained language model (of 11B parameters, which can be prohibitively large). %shows that NLE generation can be posed as a text-to-text generation task. %for a NL paradigm. 
%Similarly, 
However, NLEs are sometimes produced separately from predictions \cite{DBLP:conf/emnlp/MarasovicBPBSC20, DBLP:conf/aaai/BrahmanSRC21, DBLP:conf/acl/AtanasovaSLA20}, 
which raises questions about their faithfulness. In some cases, they were even produced as a task in isolation (without predictions) \cite{DBLP:conf/ijcnlp/JiKHWH20}. % with the task prediction process (i.e., faithfulness). 
Moreover, the majority of the existing models %\cite{DBLP:journals/corr/abs-2004-14546,kayser21,  DBLP:conf/nips/CamburuRLB18, DBLP:conf/acl/TangHS20, DBLP:journals/jmlr/RaffelSRLNMZLL20} 
only produce NLEs, with few exceptions that produce both NLEs and ERs \citep{DBLP:conf/cvpr/ParkHARSDR18, DBLP:journals/corr/abs-1809-02805}, as our model does. Furthermore, an analysis on the faithfulness of NLEs is usually missing from the large majority of these works.
% \citep{DBLP:journals/corr/abs-2004-14546, kayser21, DBLP:conf/acl/AtanasovaSLA20}. 
To our knowledge, only one work recently introduced general necessary conditions for faithfulness in NLEs \citep{DBLP:journals/corr/abs-2010-12762},
% however these conditions are still only necessary
while few other works attempted architecture-specific faithfulness measures \citep{DBLP:conf/acl/KumarT20, DBLP:journals/corr/abs-1809-02805}.

% \vspace{-1ex}
\newpara{ERs.}
%Another form of machine explanation are extractive rationales, i.e., a set of predictive input features. 
An early work \cite{DBLP:conf/emnlp/ZaidanE08} investigated rationale extraction from inputs and later was successfully followed by works for both NL \cite{DBLP:conf/acl/DeYoungJRLXSW20, DBLP:conf/emnlp/LeiBJ16, DBLP:conf/acl/BastingsAT19, LEI21AAAI} and VL \cite{DBLP:journals/corr/abs-1905-13714} tasks. %Most often, self-explainable models with ERs do not require supervision on these rationales (as it is the case in our work). 
We model both ERs and NLEs jointly in a novel framework that improves the quality of both types of explanations.

% \vspace{-1ex}
\newpara{Knowledge Grounding.}
Free-text generation tasks heavily rely on background knowledge (e.g.,~commonsense). % about the input. 
Several tasks such as dialog generation \cite{DBLP:conf/emnlp/MajumderJBM20}, creative text generation \cite{DBLP:conf/acl/ChakrabartyGMP20,DBLP:conf/emnlp/MaoMMC19}, and counterfactual generation \cite{DBLP:conf/iclr/BhagavatulaBMSH20} used commonsense for grounding. Recently, \citet{DBLP:conf/emnlp/MarasovicBPBSC20,  DBLP:conf/aaai/BrahmanSRC21} showed that external knowledge can be useful in separately justifying predictions using NLEs. In this work, we establish that knowledge grounding can be useful in a self-rationalizing framework benefiting both predictions and explanations.
% used commonsense resources as semantic understanding tools to connect low-level features to NLEs in the VL domain. 
% In this work, we establish that knowledge-grounding can be beneficial for both prediction and explainability.  %commonsense plays a critical role in connecting extractive rationales and NLEs, and help to achieve NLEs that are more grounded in commonsense (Fig.~\ref{fig:error}). 
%a new SOTA performance in explanation (both extractive and abstractive) generation across five different tasks, as well as new SOTA in task performance for two tasks (ComVE and SNLI-VE).    

%% file: files/conclusion.tex
\vspace{-1ex}
\section{Summary and Outlook}
In this work, we proposed \ours{}, a self-rationalizing framework that incorporates background knowledge resources and provides two complementary types of explanations: ERs and NLEs. Using five tasks, from natural language and vision-language domains, we show that \ours{}~obtains a new SOTA performance for both NLEs and ERs. We also close the important gap between task performance and explainability for the five tasks that we experimented with, and obtained a new SOTA for e-SNLI-VE. While we used commonsense resources, future work could look into adding other types of knowledge resources, including more specialized ones, such as legal and medical.
Additionally, while we showed that \ours{}~opens up a promising direction for zero-shot NLE generation, further investigation could reap more benefits from the principals behind \ours{}~for zero-shot and few-shot setups.

\section*{Acknowledgments}
We thank Vered Shwartz, Ana Marasović, the anonymous reviewers and meta-reviewers for their useful comments.
Bodhisattwa Prasad Majumder was partly supported by a Qualcomm Innovation Fellowship (2020), UC San Diego Friends of International Center Fellowship (2022), Adobe Research Fellowship (2022), MeetElise, and NSF Award \#1750063.
Thomas Lukasiewicz and Oana-Maria Camburu  were supported by the Alan Turing Institute under the UKRI EPSRC
grant EP/N510129/1 and by  
the UKRI EPSRC
grant EP/R013667/1. Thomas Lukasiewicz was additionally supported by 
the AXA Research Fund and by the ESRC
grant “Unlocking the Potential of AI for English Law”.

%% file: files/appendix.tex
% \appendix
% \onecolumn

% \section*{\textcolor{red}{Errata}} We have made \textbf{few typographical errors} for the placement of the decimal point in numbers in \Cref{tab:nl}. We apologize for this mistake. We urge reviewers to see \Cref{tab:all} for the true numbers. We will update these in the next version of our manuscript. 

\section{Implementation Details}
\label{sec:ap_exp}
\paragraph{Training.} We trained each model for maximum 5 epochs,  
% average
and training was stopped using an early stopping criteria based on perplexity on the validation sets.
For NL tasks, each model is trained with batch size of 4 on two 2080 Ti GPUs. Each \ours{}~variant took 35 hours on ComVE, 45 hours on e-SNLI, and 25 hours on COSe.
For VL tasks, each model is trained with batch size of 32 on two 2080 Ti GPUs. Each \ours{}~variant took 85 hours on e-SNLI-VE  and 105 hours on VCR.

\paragraph{Hyperparameters.} 
For the rationale extraction step, we set both $\lambda_0^r$ and $\lambda_1^r$ to 1.0. This value turned out to be best for both NL and VL tasks. 
For the knowledge selection step, we set $\lambda_0^g$ to 0.9, based on validation performance. The $\alpha$ for mixing rationale extraction and NLE generation loss is set to 0.4.
We use the AdamW optimizer \cite{DBLP:journals/corr/abs-1711-05101} for training each model, and the learning rate was set to $6.25e-5$, with a linear decay of step size $10^{-1}$ per epoch.
We use BART,\footnote{\url{https://huggingface.co/transformers/model_doc/bart.html}} UNITER,\footnote{\url{https://github.com/ChenRocks/UNITER}} and GPT-2,\footnote{\url{https://huggingface.co/transformers/model_doc/gpt2.html}} with all three being released under the MIT license.

\paragraph{Baselines.}
We used the official code base for NILE.\footnote{\url{https://github.com/SawanKumar28/nile}} For WT5, we fine-tuned a pretrained T5 model.\footnote{\url{https://huggingface.co/transformers/model_doc/t5.html}} For all VL baselines (PJ-X, FME, RVT, and e-UG), we followed the implementations details from \cite{kayser21}.

\section{Tasks}
\label{sec:ap_data}

\newpara{Commonsense Validation.}
We use ComVE \cite{wang-etal-2019-make}, a dataset for the task of commonsense validation, where from a pair of sentences, a model needs to choose the sentence that defies commonsense (see Fig.~\ref{fig:example}). The dataset also comes with NLEs.
ComVE consists of 10000/1000/1000 samples in the train/validation/test splits. We use the \textit{BART} tokenizer\footnote{\url{https://huggingface.co/transformers/model_doc/bart.html\#barttokenizer}} to tokenize input strings. The maximum input length was set to 512. The dataset is distributed under the CC BY-SA 4.0 license.

\newpara{Natural Language Inference.}
SNLI \cite{DBLP:conf/emnlp/BowmanAPM15} is a dataset for the task of recognizing textual entailment, where given a pair of sentences (premise and hypothesis), a model must classify their relation as either entailment, contradiction, or neutral.
% the premise entailing the hypothesis, (b) contradiction: the hypothesis contradicting the premise, or (c) neutral: neither entailment nor contradiction hold.
We use the e-SNLI \cite{DBLP:conf/nips/CamburuRLB18} dataset that contains
% 570K examples (550K/10K/1K for train/validation/test splits) along with 
NLEs for SNLI (see Fig.~\ref{fig:example}). e-SNLI consists of 550K/10K/10K samples in the train/validation/test splits. We again use the \textit{BART} tokenizer for the input strings. The maximum input length was set to 512. The dataset is distributed under the MIT license.

\newpara{Commonsense QA.}
CQA \cite{DBLP:conf/naacl/TalmorHLB19} is a multiple-choice commonsense question-answering (QA) dataset. %, where a model must use commonsense knowledge for a correct prediction. 
COSe \cite{DBLP:conf/acl/RajaniMXS19} is an extension of CQA that provides an NLE for each correct answer. 
% with 9741/1221 train/validation examples.
We treat QA as a multi-class classification
% \cite{}
task along with generating NLEs for the answer prediction.
COSe consists of 9741/1221 samples in the train/validation splits. We use the version 1.11 of the dataset. We use the \textit{BART} tokenizer to tokenize input strings. The maximum input length was set to 1024. The dataset is distributed under the BSD 3-Clause ``New'' or ``Revised'' license.

\newpara{Visual Entailment.}
SNLI-VE \cite{xie2019visual} is a vision dataset analog to the SNLI dataset \cite{DBLP:conf/emnlp/BowmanAPM15}. SNLI-VE considers an image as a premise (instead of text as in SNLI) and text as a hypothesis, with the same three labels of entailment, neutral, and contradiction. e-SNLI-VE \cite{kayser21} extends SNLI-VE with NLEs. e-SNLI-VE consists of 401K/14K/14K samples in train/validation/test splits. We use the \textit{BERT} tokenization scheme\footnote{\url{https://huggingface.co/transformers/model_doc/bert.html\#berttokenizer}} to tokenize text input following UNITER \cite{DBLP:conf/eccv/ChenLYK0G0020}. The maximum input length was set to 512. No specific license is associated with the dataset release, and the dataset is freely available.

\newpara{Visual Commonsense Reasoning.}
VCR \cite{DBLP:conf/cvpr/ZellersBFC19} is a dataset for commonsense reasoning in a visual-question-answering setup. % where the question is asked regarding an event indicated in an input image. 
%A model must use commonsense knowledge and reason about the question and the image context to answer the question.
% Along with predicting the answer, the model also needs to pick a correct commonsense explanation out of 4 explanations provided, so that the chosen one best explains the answer.
% In addition to predicting the answer, 
We generate the NLEs for each answer prediction from scratch (instead of choosing an NLE from a pool of choices, as the dataset was introduced). VCR consists of 212K/26K/26K samples in train/validation/test splits. Similar to e-SNLI-VE, we use the \textit{BERT} tokenization scheme to tokenize the input text. % following UNITER. 
The maximum input length was set to 512. The license of this dataset is mentioned at \url{https://visualcommonsense.com/license/}.

\section{Automatic Metrics}
\label{sec:ap_auto}
Following \cite{kayser21}, we experiment with a suite of metrics popularly used in language generation to capture how closely the generated NLEs follow the ground truth. We provide additional metrics that were reported in \cite{kayser21}, i.e.,
BLEU-4 \cite{papineni2002bleu}, ROUGE-L \cite{DBLP:conf/acl/LinO04}, SPICE \cite{DBLP:conf/eccv/AndersonFJG16}, CIDER \cite{DBLP:conf/cvpr/VedantamZP15} in \Cref{tab:all}. 

\begin{table}[t!]
\caption{\small \textbf{More Automatic metrics for NL and VL tasks.} Best numbers are in \textbf{bold} ($p < 0.001$).
}
\label{tab:all}
\vskip 0.08in
% \footnotesize
\centering
\small
\resizebox{\linewidth}{!}{
\centering
\begin{tabular}{ll|cccc}
\toprule
& \bf System  & \multicolumn{1}{c}{\bf BLEU} & \multicolumn{1}{c}{\bf ROUGE}  & \multicolumn{1}{c}{\bf SPICE}  & \multicolumn{1}{c}{\bf CIDER} \\ 
\midrule
% & Gold  &  -- & -- & -- & -- &  79.3 &  17.3 & 2.3 & 1.1\\
% & Neg-Heu  &  17.8 &  12.3 &  78.2 &  1.2 & 17.3 & 29.2 & 21.4 &  87.7 & 5.6 & 5.3 & 1.4\\ 
& WT5   & 21.8 &  17.2 & 24.9 & 34.1
\\ 
\rowcolor{Gray}\cellcolor{white}&\ours{}-ZS  &  14.5 &  20.4 & 16.3 & 29.4\\
\rowcolor{Gray}\cellcolor{white}&\ours{}-RB  &  23.6 &  19.8 & 27.3 & 33.5\\ 
\rowcolor{Gray}\cellcolor{white}& \ours{} w/o KN-Sel  &  22.1 &  18.5 & 24.7 & 32.3 \\
\rowcolor{Gray}\cellcolor{white} & \ours{} w/o KN \& ER  &  21.7 &  18.2 &  24.3 & 31.5 \\
\rowcolor{Gray}\cellcolor{white} \parbox[t]{0pt}{\multirow{-6}{*}{\rotatebox[origin=c]{90}{\bf ComVE}}} & \ours{}  &  \bf 25.6 &  \bf 24.5 & \bf 29.3 & \bf 37.1\\ 
\midrule
& NILE  &  29.8 &  24.3 & 34.3 & 47.4\\ 
& WT5  &  32.4 & 25.3 & 37.3 & 48.3 \\ 
\rowcolor{Gray}\cellcolor{white}&\ours{}-ZS  &  25.5 &  24.4 & 33.6 & 40.1\\
\rowcolor{Gray}\cellcolor{white}&\ours{}-RB  &  35.6 &  28.9 & 39.8 & 52.5\\ 
\rowcolor{Gray}\cellcolor{white}& \ours{} w/o KN-Sel  &  30.5 &  24.9 & 35.8 & 47.3 \\
\rowcolor{Gray}\cellcolor{white}& \ours{} w/o KN \& ER  &  31.5 &  25.3 & 36.4 & 48.3 \\
\rowcolor{Gray}\cellcolor{white}\parbox[t]{2mm}{\multirow{-7}{*}{\rotatebox[origin=c]{90}{\bf e-SNLI}}} &\ours{}  &  \bf 37.9 &  \bf 32.4 & \bf 42.6 & \bf 54.4\\
\midrule 
& GPT-2  & 1.3  & 8.2 & 13.2 & 22.3 \\ 
& WT5  & 4.2 &  12.3 & 18.3 & 27.2 \\
\rowcolor{Gray}\cellcolor{white}& \ours{}-ZS  &  3.2 &  9.4 & 15.4 & 23.1 \\  
\rowcolor{Gray}\cellcolor{white}& \ours{}-RB  &  3.8 &  9.8 & 16.9 & 27.9 \\
\rowcolor{Gray}\cellcolor{white}& \ours{} w/o KN-Sel  &  4.1 &  10.9 & 17.1 & 26.9 \\  
\rowcolor{Gray}\cellcolor{white}& \ours{} w/o KN \& ER  &  4.2 &  11.4 & 17.3 & 27.4 \\  
\rowcolor{Gray}\cellcolor{white}\parbox[t]{2mm}{\multirow{-7}{*}{\rotatebox[origin=c]{90}{\bf COSe}}} & \ours{}  &  \bf 5.5 &  \bf 18.3 & \bf 24.3 & \bf 35.4 \\ 
% \midrule
\midrule
& PJ-X  & 7.3 & 28.6  &  24.3 & 72.5 \\
& FME  & 8.2  & 29.9  & 26.8 & 83.6 \\
& RVT  &  9.6 &  27.3  & 32.5 & 81.7 \\ 
& e-UG & 9.6 & 27.8 & 34.5 & 85.9 \\
\rowcolor{Gray}\cellcolor{white}& \ours{}-ZS  &  7.6 &  24.1 & 33.2 & 80.3 \\  
\rowcolor{Gray}\cellcolor{white}& \ours{}-RB  &  9.8 &  26.6 & 35.1 & 86.0 \\
\rowcolor{Gray}\cellcolor{white}& \ours{} w/o KN-Sel  &  10.9 &  27.8 & 35.9 & 87.2 \\  
\rowcolor{Gray}\cellcolor{white}& \ours{} w/o KN \& ER  &  10.1 &  27.4 & 35.3 & 86.1 \\
\rowcolor{Gray}\cellcolor{white}\parbox[t]{2mm}{\multirow{-9}{*}{\rotatebox[origin=c]{90}{\bf e-SNLI-VE}}} & \ours{}  &  \bf 11.2 &  \bf 28.5  & \bf 36.9 & \bf 88.2 \\ 
\midrule
& PJ-X  & 3.4  & 20.5 & 4.5 & 19.0 \\
& FME  & 4.4  & 22.7 & 24.2 & 27.7 \\ 
& RVT  &  3.8 &  21.9 & 11.7 & 30.1 \\ 
& e-UG & 4.3 & 22.5 & 12.6 & 32.7 \\
\rowcolor{Gray}\cellcolor{white}& \ours{}-ZS  &  3.6 &  22.1 & 25.9 & 25.6 \\  
\rowcolor{Gray}\cellcolor{white}& \ours{}-RB  &  4.9 &  24.9 & 28.4 & 28.2 \\
\rowcolor{Gray}\cellcolor{white}& \ours{} w/o KN-Sel  &  5.3 &  24.8 & 28.5 & 28.3 \\  
\rowcolor{Gray}\cellcolor{white}& \ours{} w/o KN \& ER  &  5.1 &  24.4 & 28.2 & 27.9 \\
\rowcolor{Gray}\cellcolor{white}\parbox[t]{2mm}{\multirow{-9}{*}{\rotatebox[origin=c]{90}{\bf VCR}}} & \ours{}   &  \bf 5.9 &  \bf 25.4 & \bf 29.1 & \bf 29.8 \\
\bottomrule
\end{tabular}
}
\end{table}

\section{Human Evaluation}
\label{sec:ap_human}
\begin{figure*}[t!]
    \centering
    \includegraphics[width=0.8\textwidth]{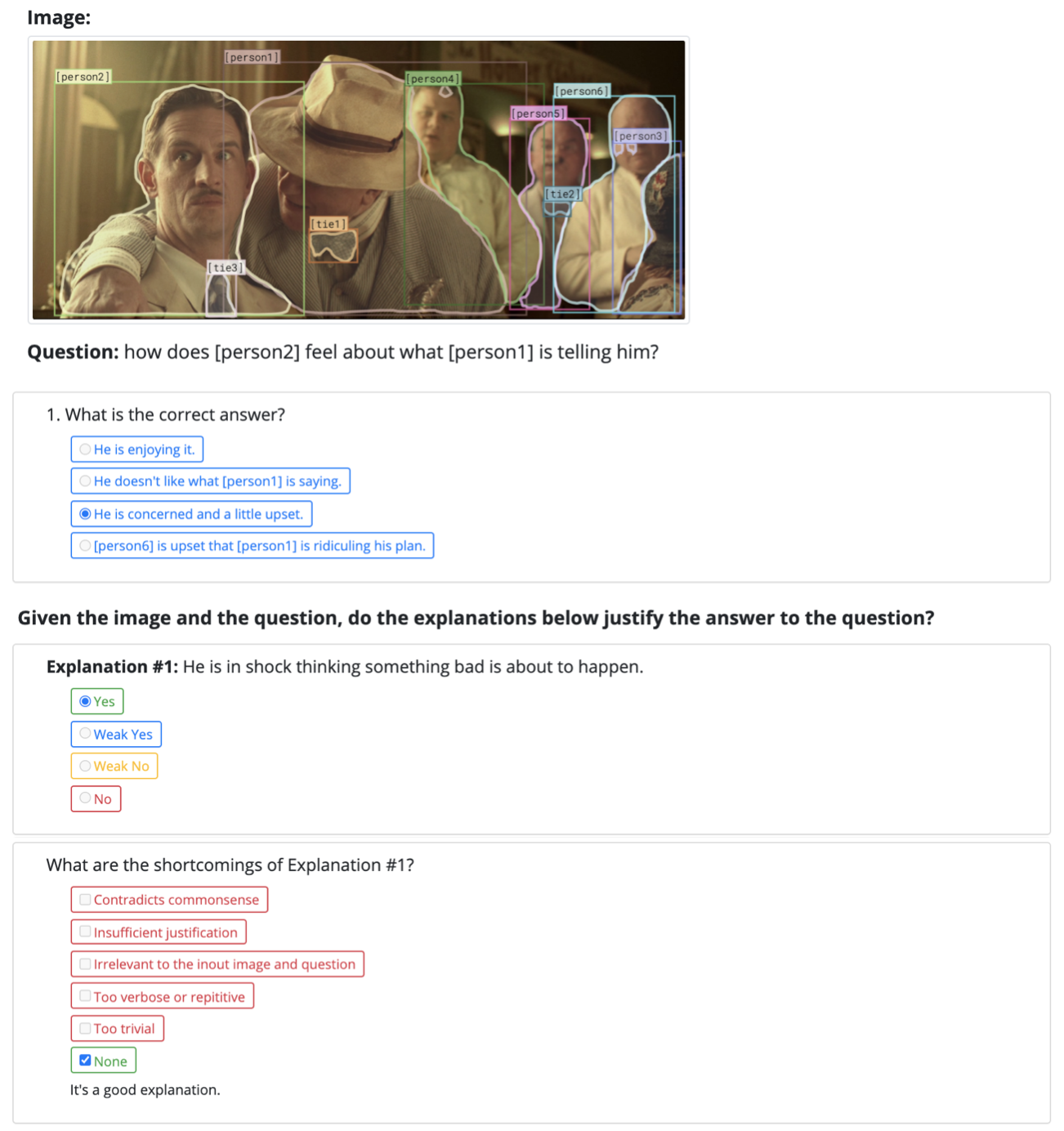}
    \caption{\small Snapshot of our human evaluation with a list of possible shortcomings.} 
    \label{fig:human_ui}
    \vspace{-1em}
\end{figure*}

We designed the human evaluation study based on \citep{kayser21} to assess the NLE quality using Amazon Mechanical Turk.
% We asked for two different Anglophone (Lifetime HIT acceptance \% $>$ 85) annotators for every test sample.
% showed that none of the automatic metrics consistently show correlation with human evaluation. Hence, we performed human evaluation for NLEs and we provide the detailed results in \Cref{tab:all}. 
We briefly describe the human evaluation setup here, with a representative snapshot of the UI shown in Fig.~\ref{fig:human_ui}. For every question, we employed two Anglophone annotators with lifetime HIT acceptance rate of at least 90\%. 

We made sure that the human annotators are able to solve the predictive task before they evaluate the NLEs.
For each NLE, we ask: \textit{Given the input, does the explanation justify the answer?} and provide four options: Yes, Weak-Yes, Weak-No, and No. We report the e-ViL score from \cite{kayser21} combining results for each option. We only consider NLEs for correct predictions and collect 250 random such examples for each model and each dataset. The inter-annotator agreement was captured by Cohen's Kappa \cite{cohen1960coefficient}. For each of the datasets, ComVE, e-SNLI, COSe, e-SNLI-VE, and VCR, the inter-annotator agreement (kappa) was 0.72, 0.76, 0.79, 0.81, and 0.74, respectively.

\begin{figure}[t!]
    \centering
    \includegraphics[trim=210 190 545 190,clip, width=\linewidth]{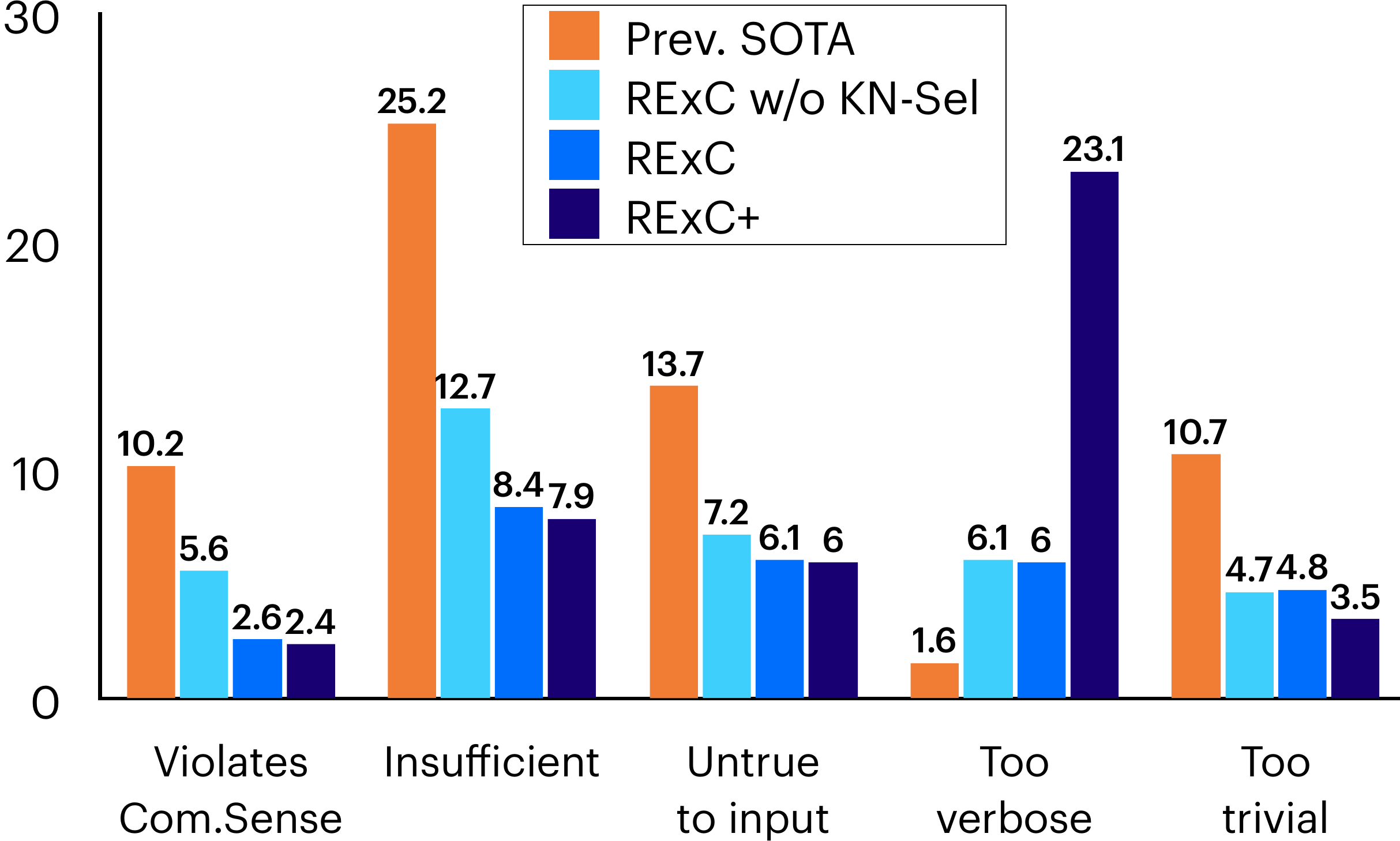}
    \caption{\small \textbf{Main limitations of the generated NLEs} obtained from user study. All numbers are in \% and are averaged by systems and datasets for both NL and VL tasks. Human annotators could choose multiple limitations for an NLE.
    % Neg-Heu and NILE are reported only for ComVE and e-SNLI, respectively.
    }
    \label{fig:error}
    % \vspace{-1.5em}
\end{figure}

\newpara{Error analysis.} \Cref{fig:error} summarizes the main drawbacks of generated NLEs (in average) across models and datasets.
% (in average)
As main observation, we see that adding commonsense knowledge and knowledge selection in \ours{}~gradually make NLEs more comprehensive and more relevant to the input. While \ours{}+ wins over all other models across all datasets, human judges often found them too verbose due the presence of supporting knowledge snippets, which might repeat information from the generated NLEs.

\begin{figure*}[h!]
    \centering
    \includegraphics[trim= 65 125 230 75,clip, width=0.8\textwidth]{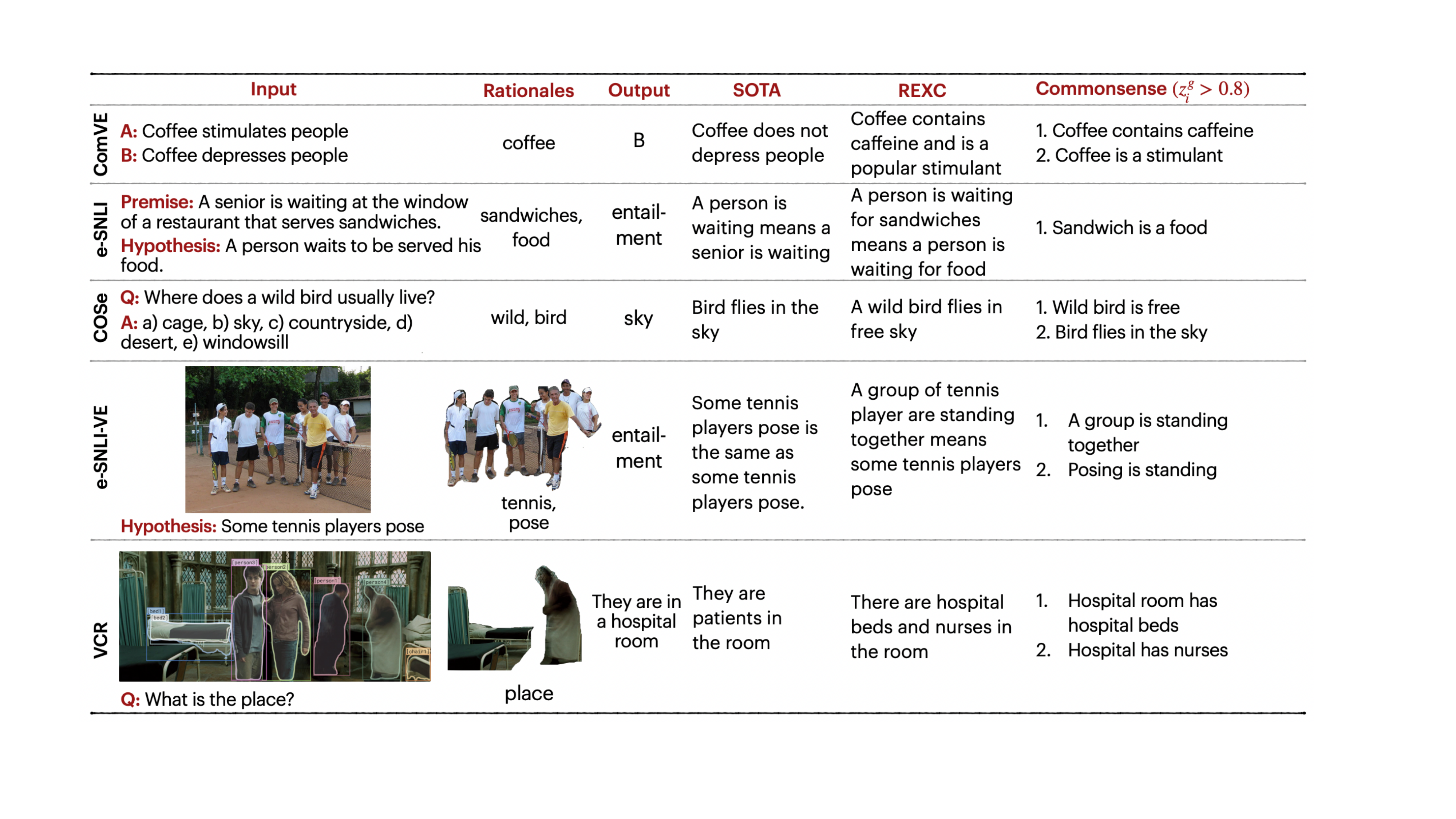}
    \caption{\small Examples of NLEs and extractive rationales generated from \ours{} for all five tasks, along with the pieces of commonsense used by \ours{}. Generations from the best baseline are included for direct comparison.} 
    \label{fig:example2}
    \vspace{-0.5em}
\end{figure*}

Another set of illustrative examples is also given in Fig.~\ref{fig:example2}. 
% ---
% As expected, for ComVE, the naive baseline Neg-Heu turns out to produce highly trivial NLEs, even if they get better eViL scores than most baselines due to their construction.
% more relevant and commonsense-grounded (by construction).  

\section{Faithfulness}
\label{sec:ap_faith}
\begin{figure*}[t!]
    \centering
    \includegraphics[width=0.9\linewidth]{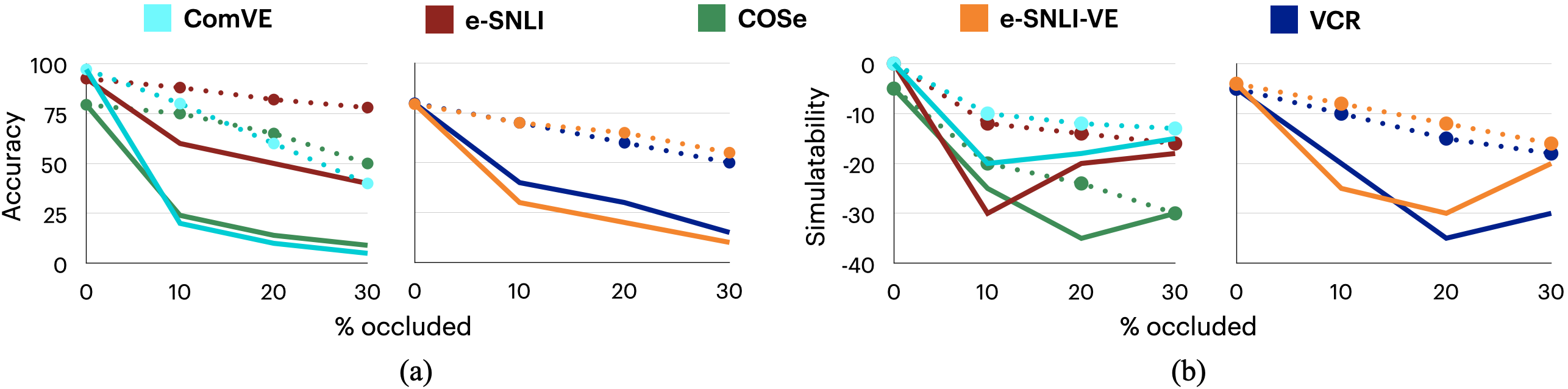}
    \vspace{-2ex}
    \caption{\small \textbf{Feature importance agreement} with (a) task accuracy and (b) NLE simulatability for all tasks. Details in \Cref{sec:faith}.
    % (a) Solid lines indicate the accuracy for predicted labels when features important for NLEs are occluded. The dotted lines indicate the prediction accuracy when random tokens (or super-pixels) are dropped. (b) Solid lines indicate the simulatabilities when features important for prediction are occluded. Dotted lines indicates simulatabilities for random occlusions. Solid lines should be lower (means higher changes) than dotted lines for better label-NLE association.
    % % For better label-NLE association, the solid lines both in (c) and (d) should have lower values than the dotted lines.
    % selected by \ours{}. 
    % Generations from the best baseline are included for direct comparison. Both SOTA and \ours{}~produce correct prediction, as shown here.
    } 
    \label{fig:feature_agreement_all}
\end{figure*}

\begin{figure*}[t!]
    \centering
    \includegraphics[width=0.7\textwidth]{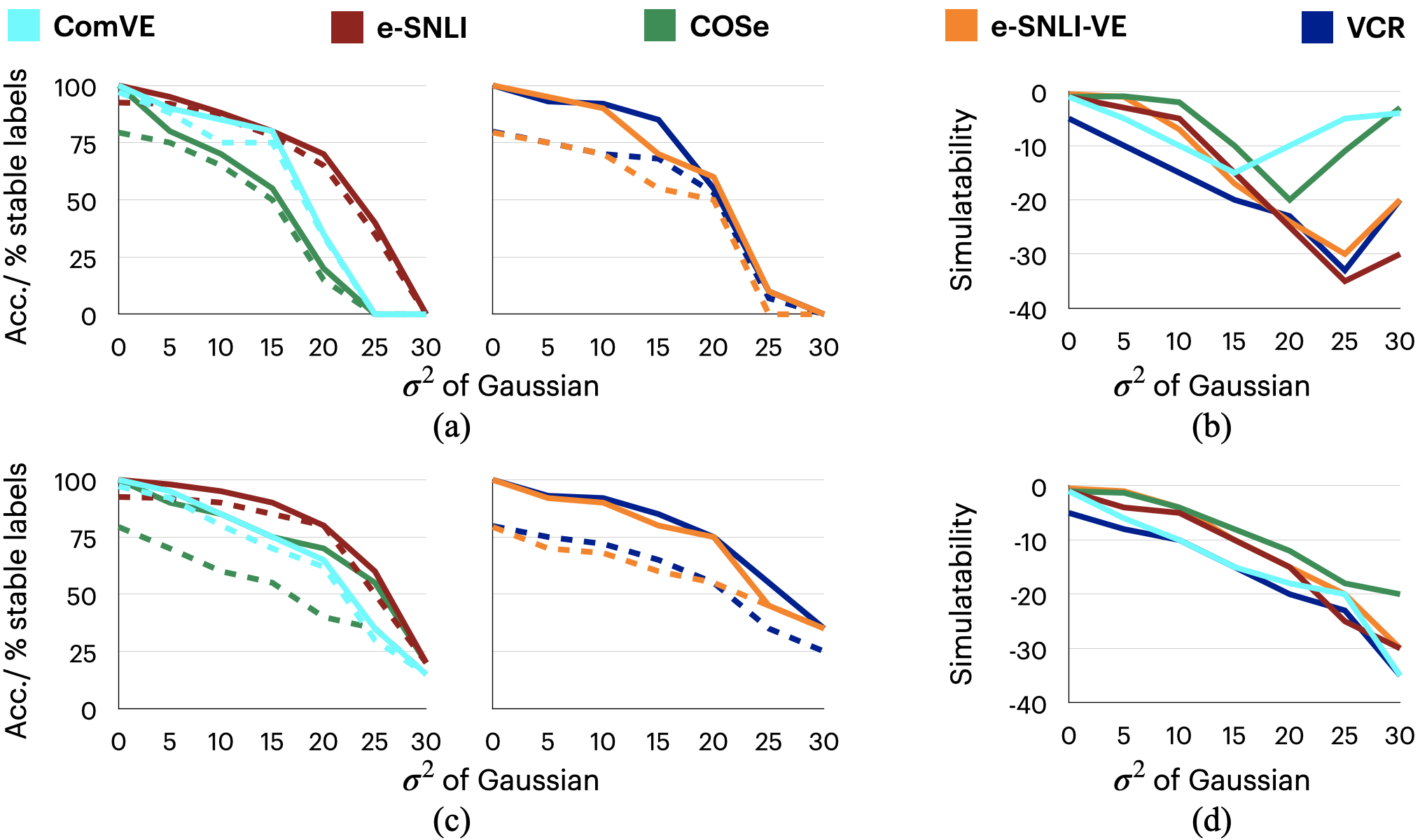}
    \vspace{-1em}
    \caption{\small \textbf{Robustness equivalence} analysis when noise (with various $\sigma^2$) is added in (a, b) input and (c, d) selected knowledge snippets for all tasks. Details in \Cref{sec:faith}.
    } 
    \label{fig:faith-noise-all}
    \vspace{-0.5em}
\end{figure*}

For all datasets, we observe feature importance agreement between labels and NLE, as shown in Fig.~\ref{fig:feature_agreement_all}. Similarly, we see that labels and NLEs are equivalently robust for all datasets, as shown in Fig.~\ref{fig:faith-noise-all}. This confirms that there exists a strong label-NLE association for \ours{}---satisfying the necessary conditions for faithful explanations.

\vspace{-2ex}
\vspace{-2ex}

% \section{More Ablations}

% % \newpara{More Ablation  \ours{}.}
% To pinpoint the contributing factor behind the improved performance in NLE generation and in the predictive task (\Cref{tab:nl,tab:vl}), we individually drop modules responsible for rationale extraction and knowleddge selection and study relative performance changes. For both NL and VL tasks, we find that an ablative baseline (\ours{} w/o KN-Sel) that does not use external knowledge resources consistently lags in NLE quality compared to 
% % all variants of 
% \ours{}. Similarly, in \autoref{tab:task}, we observe performance drops in all prediction tasks (e.g.,~2.7 and 2.5 points in VCR) when we ablate with commonsense and rationale extraction, respectively. Without knowledge selection, the performance of \ours{}~drops both for NLEs and task predictions. 
% Hence,  (1) NLEs (and not just the %external
% commonsense) can enhance a model's predictive performance and (2) knowledge selection acts as a useful inductive bias in effectively utilizing a large pool of knowledge.

